\documentclass[pre,preprint,showpacs,amsmath,amssymb,floatfix]{revtex4-1}

\usepackage{setspace}
\usepackage{multirow}
\usepackage{latexsym}
\usepackage{amssymb,amsmath,amsfonts}
\usepackage{bm,bbm}
\usepackage{graphicx,epsfig}
\usepackage{caption}
\usepackage{subcaption}
\usepackage{url}

\newcommand{\bftheta}{{\bm \theta}}

\newcommand{\bfs}{{\bf s}}

\newcommand{\bfr}{{\bf r}}
\newcommand{\bfh}{{\bf h}}
\newcommand{\bfz}{{\bf s}}

\newcommand{\al}{\alpha}

\newcommand{\e}{\mathrm{e}}
\newcommand{\E}{\mathbb{E}}
\newcommand{\R}{\mathbb{R}}

\newcommand{\Xo}{x}
\newcommand{\bXo}{\mathbf{x}}
\newcommand{\bXn}{\bXo_{n}}

\newcommand{\hXo}{\hat{x}}
\newcommand{\bhXo}{\mathbf{\hat{x}}}

\newcommand{\w}{u_{i,j}}

\newcommand{\xsam}{\bXo_{\mathrm S}}
\newcommand{\xpre}{\bhXo_{\mathrm P}}

\newcommand{\bmx}{\bm{\mu}_{\mathrm X}}
\newcommand{\mx}{\mu_{\mathrm X}}
\newcommand{\smx}{\overline{\mu}_{\mathrm X}}

\newcommand{\mcal}{\mathcal}

\newcommand{\om}{\omega}

\newcommand{\Hx}{H_{\rm X}}
\newcommand{\hHx}{\hat{H}_{\rm X}}
\newcommand{\Hconv}{{\mathcal H}(S_{N})}

\newcommand{\Sn}{\textsl{S}_{\rm N}}
\newcommand{\bmthe}{{\bm \theta}}

\newcommand{\beq}{\begin{equation}}
\newcommand{\eeq}{\end{equation}}

\newcommand{\cone}{d}

\begin{document}

\title{Stochastic Local Interaction (SLI) Model: Interfacing Machine Learning and Geostatistics}

\author{Dionissios T. Hristopulos}
 \email{dionisi@mred.tuc.gr}   
 \affiliation{School of Mineral Resources Engineering, Technical University of Crete,
Chania 73100, Greece}

\date{\today}

\begin{abstract}
Machine learning and geostatistics are powerful mathematical frameworks for modeling spatial data.
Both approaches, however, suffer from poor scaling of the required computational resources for large data applications.
We present  the Stochastic Local Interaction (SLI) model, which employs a
local representation to improve computational efficiency.
SLI combines geostatistics and machine learning with ideas from
 statistical physics and computational geometry.
It is based on a joint probability density
function defined by an energy functional which involves
local interactions implemented by means of kernel functions with  adaptive local kernel bandwidths.
SLI is expressed in terms of an explicit, typically sparse,
 precision (inverse covariance) matrix.
This representation leads to a semi-analytical  expression for interpolation (prediction), which is
valid in any number of dimensions and avoids the computationally costly covariance matrix inversion.
\end{abstract}



\maketitle

\newpage

\section{Introduction}

Big data is expected to have a large impact in the
geosciences given the abundance of remote sensing and earth-based observations related to climate~\cite{Steed13,Addair14}.
A similar data explosion is happening
in other scientific and engineering fields~\cite{Wu14}.
This trend underscores the need for algorithms that can handle large data sets.
Most current methods of data analysis, however, have not been designed with size as a primary consideration.
This has inspired statements such as: ``Improvements in data-processing capabilities are essential
to make maximal use of state-of-the-art experimental facilities''~\cite{Ahrens11}.
Machine learning  can extract information and
``learn'' characteristic patterns in the data. Thus, it is expected to play a significant
role in the era of big data research. The application
of machine learning methods in spatial data analysis has been spearheaded by Kanevski~\cite{Kanevski04}.
Machine learning and geostatistics  are powerful frameworks for spatial data processing.
A comparison of their performance using a set of radiological measurements is presented in~\cite{Giralids05}.
The question that we address in this work is whether we can combine ideas from both fields to
develop a computationally efficient framework for spatial data modeling.

Most data processing and visualization methods assume complete data sets,
whereas in practice data often have gaps.
Hence, it is necessary to fill missing values by means of imputation or interpolation methods.
In geostatistics, such methods are based on various flavors of stochastic optimal linear estimation
(kriging)~\cite{Chiles12}. In machine learning, methods such as $k$-nearest neighbors,
 artificial neural networks, and the Bayesian framework of Gaussian processes are used~\cite{Rasmussen06}.
Both geostatistics and Gaussian process regression  are based on the theory of random fields and share considerable similarities~\cite{Adler81,Yaglom87}.
The Gaussian process framework, however, is better suited for applications in higher than two dimensions.
A significant drawback of most existing methods for interpolation and simulation of missing data
is their poor scalability with the data size $N$, i.e., the  $O(N^3)$ algorithmic complexity and the $O(N^2)$ memory requirements:
An $O(N^p)$ dependence  implies that the respective computational resource   (time or memory) increases
with $N$ as a polynomial of degree at most equal to $p$.

Improved scaling with data size can be achieved by means of local approximations,
dimensionality reduction techniques, and parallel algorithms.
A recent review of available methods for
large data geostatistical  applications is given  in~\cite{Sun12}.
These approaches employ clever approximations to reduce the computational complexity
of the standard geostatistical framework. Local approximations involve methods such as
maximum composite likelihood~\cite{Varin11} and  maximum pseudo-likelihood~\cite{VanDuijn09}.
Another approach involves covariance tapering which neglects correlations outside a specified range~\cite{Furrer06,Kaufman08,Du09}.
Dimensionality reduction includes methods such as
fixed rank kriging which models the precision matrix by means of  a fixed rank  matrix $r \ll N$~\cite{Cressie08,Nychka14}.
Markov random fields (MRFs)  also take advantage of  locality using factorizable joint densities.
The application of MRFs in spatial data analysis was initially limited to
structured grids~\cite{Rue05}. However, a recently developed link between Gaussian random fields
and MRFs via stochastic partial differential equations (SPDE) has extended the scope  of MRFs
to scattered data~\cite{Rue10}.

We propose a Stochastic Local Interaction (SLI) model for  spatially correlated data
which is based by construction on local correlations.
SLI can be used for the
interpolation and simulation  of incomplete data in $d$-dimensional spaces, where  $d$ could be
larger than $3$.
The SLI model incorporates concepts from statistical physics, computational geometry, and machine learning.
We use the idea of local interactions from statistical physics to impose correlations between
``neighboring'' locations by means of an explicit precision matrix.
The local geometry of the sampling network plays an important role
in the expression of the interactions, since it determines
the size of local neighborhoods.
On regular grids, the SLI model becomes equivalent to a Gaussian MRF with specific
structure. For scattered data, the SLI model provides an alternative to the SPDE
approach that avoids the preprocessing cost involved in the latter.

The SLI model extends previous research on Spartan spatial
 random fields~\cite{dth03b,dthsel07,dthetal08} to an explicitly discrete formulation and
 thus enables its application to scattered data without the approximations used in~\cite{dthsel09}.
 SLI is based on a joint probability density
function (pdf) determined from local interactions.
This is achieved by handling the irregularity of sampling locations in terms of
 kernel functions with locally adaptive bandwidth.
Kernel methods are common in statistical machine
learning~\citep{Vapnik00} and in spatial statistics for the estimation of
 the variogram and the covariance function~\citep{Elogneetal07,Garcia04,Hall94}.

The remainder of the article is structured as follows:
Section~\ref{sec:back}  briefly introduces useful definitions and terminology.
In Section~\ref{sec:method} we construct the SLI model, propose a computationally efficient
parameter estimation approach, and
formulate an explicit interpolation expression.
 In Section~\ref{sec:case}
we investigate SLI interpolation using different types of simulated and real data in
one, two and four dimensional Euclidean spaces.
Section~\ref{sec:disc}
discusses potential extensions of the current SLI version and
connections with machine learning. Finally, in Section~\ref{sec:concl} we present our conclusions and
point to future research.

\section{Background Concepts and Notation}
\label{sec:back}

\subsection{Definition of the problem to be learned}

\paragraph{Sampling grid} The set of sampling points is denoted by $\Sn=\{\bfs_{1},\ldots, \bfs_{N}\}$,
where $\bfs_{i},$ $i=1,\ldots,N$ are vectors
in the Euclidean space $\R^d$ or in some abstract feature space that possesses a distance metric.
 In Euclidean spaces, the domain boundary is defined by the convex
hull, $\Hconv,$ of $\Sn.$

\paragraph{Sample and predictions} The sample data are denoted by the vector $\xsam \equiv
(\Xo_{1}, \ldots, \Xo_{N})^{T}$, where the superscript ``$T$'' denotes
the transpose. Interpolation aims to derive  estimates
of the observed field at the nodes of a regular grid
$\mathcal{G} \subset {\mathbb Z}^{d}$, or at validation set  points
which may be scattered.
The estimates (predictions) will be denoted by
$\hXo(\bfz_p)$, $p=1,\ldots,P$, i.e., $\xpre =(\hXo_1, \ldots, \hXo_{P})^{T}$.

\paragraph{Spatial random field model} The data $\xsam$  are assumed to represent samples from
a spatial random field (SRF) $X_{i}(\om)$,
 where the index $i=1, \ldots, N$ denotes the spatial location $\bfs_i \in \Sn$.
 The expectation over the ensemble of probable states is denoted by
 $\E[X_{i}(\om)]$, and the autocovariance function is given by
  $C_{i,j} := \E[X_{i}(\om)\, X_{j}(\om)] - \E[X_{i}(\om)]\, \E[X_{j}(\om)].$

The pdf of \textit{Gibbs SRFs} can be
expressed in terms of an energy functional
$ H(\xsam;\bmthe) $, where $\bmthe$ is a set
of \textit{model parameters}, according to the  Gibbs pdf
\citep[p. 51]{Winkler95}

\begin{equation}
\label{eq:gibbspdf}
f_{\rm X} (\xsam;\bmthe) =
\frac{\e^{ - H(\xsam;\bmthe)}  }{Z(\bmthe)}.
\end{equation}

\noindent
The constant $ Z(\bmthe) $, called the \textit{partition function},
is the pdf normalization factor obtained by integrating
$ \e^{ -H(\xsam;\bmthe) } $ over all
the probable states $\xsam$.

\subsection{From continuum spaces to scattered data}
The formulation based on~\eqref{eq:gibbspdf}  has its origins in statistical physics,
and it has found applications in
pattern analysis~\cite{Besag74,Geman84} and Bayesian field theory, e.g.~\cite{Lemm05,Farmer07}.
In statistics, this general model belongs to the exponential family of distributions
that have desirable mathematical properties~\cite{Barndorff14}.
Our group used the exponential density in connection with a specific energy functional to
 develop Spartan spatial random fields (SSRF's)~\citep{dth03b,dthsel07,dthetal08,dthsel09}.
In Section~\ref{sec:method} we construct an explicitly discrete model motivated by SSRFs which
adapts local interactions to general sampling networks
and prediction grids by means of kernel functions.

\subsection{Kernel weights}
\label{ssec:k-weights}
Let $K(\bfr)$ be a non-negative-valued kernel that is either
compactly supported or decays exponentially
fast at large distances (e.g., the Gaussian or exponential function).
We define  kernel weights  associated with the sampling points $\bfs_{i}$ and
$\bfs_{j}$ as follows

\beq
K_{i,j} \doteq
K\left(\frac{\bfs_{i} - \bfs_{j}}{h_{i}}\right) = K\left(\frac{\|\bfs_{i} - \bfs_{j}\|}{h_{i}}\right),
\eeq

\noindent
where $\|\bfs_{i} - \bfs_{j}\|$ is the distance (Euclidean or other) between two points $\bfs_{i}$ and $\bfs_{j} $, whereas
$h_i$ is the respective \emph{kernel bandwidth} that adapts to local variations of
the sampling  pattern. The kernel weight $K_{j,i}$ is defined in terms of a bandwidth $h_j$.
Hence,  $K_{i,j} \neq K_{j,i}$ if the
bandwidths $h_i$  and $h_j$ are different. Examples of  kernel functions
 are given in Table~\ref{tab:kernels}.
\begin{table}
\centering
\caption{\label{tab:kernels} Definitions of kernel functions used in Section~\ref{sec:case} below.
The first three have compact support.
Notation:   $u = \| \bfr \|/h$ where $\| \bfr \| $ is the distance and
$h$ the bandwidth;   $\mathbbm{1}_{A}(u)$ is the
{indicator function}
of the set $A$, i.e., $\mathbbm{1}_{A}(u)=1, \, u \in A$
and $\mathbbm{1}_{A}(u)=0, \,u \notin A.$
} \centering
\begin{tabular}{lll}
\hline
Triangular & $K(u)=(1-u)\, \mathbbm{1}_{|u| \le 1}(u)$  \\
Tricube  & $K(u)=(1-u^3)^3 \, \mathbbm{1}_{|u| \le 1}(u)$  \\
Quadratic & $K(u)=(1-u^2) \, \mathbbm{1}_{|u| \le 1}(u)$  \\
Gaussian  & $K(u)=\exp(-u^2)$  \\
Exponential & $K(u)=\exp(-|u|)$    \\
\hline
\end{tabular}
\end{table}

Let $D_{i,[k]}(S_{N})$ denote the distance between  ${\bfs}_{i}$
and its $k$-nearest neighbor in  $\Sn$ $(k=0$ corresponds to zero distance).
We choose the local bandwidth associated with ${\bfs}_{i}$ according to

\beq
\label{eq:bandwidth}
h_{i} = \mu\, D_{i,[k]}(S_{N}),
\eeq

\noindent where $\mu > 1$ and $k>1$ are model parameters.
In several case studies involving Euclidean  spaces of dimension
$d=1, 2, 3, 4$, we determined that $k=2$ (second nearest neighbors)
performs well for compactly supported kernels and $k=1$ (nearest neighbors) for
infinitely supported kernels.
Using $k=2$ for compact kernels
 avoids zero bandwidth problems  which  result from $k=1$ for
collocated sampling and prediction points.
Since the sampling point configuration
is fixed, $\mu$ and $D_{i,[k]}(S_{N})$ determine the local bandwidths.
 $D_{i,[k]}(S_{N})$ depends purely on the sampling point configuration, but
$\mu$ also depends on the sample values.
For compactly supported kernels setting $k=1$ only makes sense if $\mu >1$;
otherwise $h_{i} = D_{i,[k=1]}(S_{N})$ implying that the kernel vanishes even for the nearest-neighbor pairs
and thus fails to implement interactions.

\subsection{Kernel averages}
For any two-point function $\Phi(\cdot)$, we use a local-bandwidth extension of the
Nadaraya-Watson kernel-weighted average over the
network of sampling points~\cite{Nadaraya64,Watson64}

\[
\langle \Phi(\cdot) \rangle_{\bfh} = \frac{\sum_{i=1}^{N} \sum_{j=1}^{N} K_{i,j} \, \Phi(\cdot)}
{\sum_{i=1}^{N} \sum_{j=1}^{N} K_{i,j} },
\]

\noindent where $\bfh = (h_{1}, \ldots, h_{N})^T$ is the vector of local bandwidths.
The function $\Phi(\cdot)$  represents the distance between two points $\|\bfs_{i} - \bfs_{j}\|$ or
the difference $x_{i} - x_{j}$ of the field values, or any other function that depends on
the locations or the values of the field. The kernel average is  normalized so as to preserve
unity, i.e., $\langle 1 \rangle_{\bfh} = 1$ for all possible point configurations.

\section{The Stochastic Local Interaction (SLI) Model}
\label{sec:method}

The  SLI joint pdf is determined below by means of the energy functional~\eqref{eq:fgc-disc}.
This leads to a precision matrix which is explicitly defined in terms  of local
interactions and thus avoids the  covariance matrix inversion.
The prediction of missing data is based on maximizing the joint pdf of the data and the predictand,
which is equivalent to minimizing the corresponding energy functional.
This leads to the mode predictor~\eqref{eq:SLI-mode-pred},
 which involves a calculation with linear algorithmic complexity.

\subsection{The energy functional}
Consider a sample $\xsam$ on an unstructured
sampling grid with sample mean $\mx$.
We propose the following energy functional
$\Hx(\xsam;\bftheta)$

\begin{align}
\label{eq:fgc-disc} \Hx (\xsam ;\bftheta) &    =
\frac{1}{2\,\lambda \, }   \left[ {\mcal{S}_0}(\xsam) + \, \alpha_1 \,
{\mcal{S}_1}(\xsam;\bfh_1) \right.
    + \left. \alpha_{2} \, {\mcal{S}_2}(\xsam;\bfh_2) \right],
\end{align}

\noindent
where $\bftheta=(\mx, \alpha_1, \alpha_2, \lambda, \mu, k)$ is the SLI parameter vector
and the parameters $\mu, k$ are defined in Section~\ref{ssec:k-weights} above.

The terms $\mcal{S}_0(\xsam)$, ${\mcal{S}_1}(\xsam;\bfh_1)$, and ${\mcal{S}_2}(\xsam;\bfh_2)$
 correspond to the averages of the
square fluctuations, the square gradient and the square curvature in a Euclidean space of
dimension $d$. The latter two
 are given by
kernel-weighted averages that involve the \emph{field increments} $\Xo_{i,j} = x_{i} - x_{j}$.

\beq
{\mcal{S}_0}(\xsam) = \frac{1}{N}\,{\sum_{i=1}^{N}
(\Xo_{i}-\mx)^{2}},
\eeq
\begin{equation}
\label{eq:S1}
{\mcal{S}_1}(\xsam;\bfh_1) = \cone  \, \langle \Xo_{i,j}^{2} \rangle_{\bfh_1},
\end{equation}
\begin{subequations}
\begin{align}
\label{eq:S2}
{\mcal{S}_2}(\xsam;\bfh_2) = c_{2,1} \, \langle \Xo_{i,j}^{2} \rangle_{\bfh_2}
 - c_{2,2} \, \langle \Xo_{i,j}^{2} \rangle_{\bfh_3}
 - c_{2,3} \, \langle \Xo_{i,j}^{2} \rangle_{\bfh_4},
\\
\mbox{where} \; c_{2,1}= 4d(d+2), \; c_{2,2} = 2d(d-1), \; c_{2,3} = d.
\end{align}
\end{subequations}

The $c_{2,j}$ $(j=1,2,3)$ values in  $\mcal{S}_2$ are motivated by
  discrete approximations
of the square gradient and curvature~\cite{dthsel09}.
We use two vector bandwidths, $\bfh_1$ and $\bfh_2$,  to determine the
 range of influence of the kernel function around each sampling point for the gradient
${\mathcal{S}_1}(\bXn; \bfh_1)$ and curvature
${\mathcal{S}_2}(\bXn; \bfh_2)$ terms respectively.
Additional bandwidths used in~\eqref{eq:S2} for ${\mathcal{S}_2}(\bXn; \bfh_2)$
are defined by $\bfh_3=\sqrt{2}\,\bfh_2$, $\bfh_4=2\,\bfh_2.$
These definitions are motivated by the formulation of SSRFs~\cite{dth03b,dthsel09}.

\subsection{SLI parameters and permissibility}
\label{ssec:sli-permissibility}
To obtain realistic kernel bandwidths,   $k$ should be a positive integer larger than one,
and $\mu $ should be larger than one.
The parameter $\mu_{\rm X}$ is set equal to the sample mean.
The coefficients $\alpha_1, \alpha_2$ control the relative contributions
of the mean square gradient and
mean square curvature terms.
The coefficient $\lambda$ controls the overall amplitude of the fluctuations. Finally,
$\mu$ and $k$ control the bandwidth values as described in Section~\ref{ssec:k-weights}.

 The SLI energy functional~\eqref{eq:fgc-disc} is permissible if $\Hx (\xsam ;\bftheta) \ge 0$
 for all $\xsam$, a condition which ensures that $ \e^{ -H(\xsam;\bmthe) } $  is bounded and
 thus the
existence of the partition function in~\eqref{eq:gibbspdf}.
Assuming that $\mcal{S}_2 \ge 0$ ($\mcal{S}_0$ and $\mcal{S}_1$ are always non-negative
by construction),
a sufficient permissibility condition, independently of the distance metric used, is
$\alpha_1, \alpha_2, \lambda >0$.
In all the case studies that we have investigated, however,
we have not encountered permissibility problems so long as
$\alpha_1, \alpha_2, \lambda >0$.
Intuitively, the justification for the permissibility of~\eqref{eq:fgc-disc} is that the first average, i.e.,
$\langle \Xo_{i,j}^{2} \rangle_{\bfh_2}$  in~\eqref{eq:S2} has a
positive sign and is multiplied by
$c_{2,1}$, which is significantly larger
(especially as $d$ increases) than the coefficients $c_{2,2}$ and $c_{2,3}$ multiplying the
negative-sign averages  $\langle \Xo_{i,j}^{2} \rangle_{\bfh_3}$  and $\langle \Xo_{i,j}^{2} \rangle_{\bfh_4}$.
This property is valid for geodesic distances on the globe and for
other metric spaces as well.


\subsection{Precision matrix representation}
\label{ssec:prec-mat}

We express~\eqref{eq:fgc-disc}  in terms
of the \emph{precision matrix} $\hat{J}_{i,j}(\bftheta)$ ($i,j =1, \ldots, N$)

\begin{equation}
\label{eq:H-using-J} \Hx (\xsam ;\bftheta)  =
\frac{1}{2}
  (\xsam - \bmx)^{T} \, {\mathbf J}(\bftheta) \,(\xsam - \bmx).
\end{equation}

The symmetric precision matrix ${\mathbf J}(\bftheta)$ follows from expanding the squared differences
 in~\eqref{eq:fgc-disc}, leading to the following expression

 \begin{align}
\label{eq:prec-mat} {\mathbf J}(\bftheta)    =     \frac{1}{\lambda } &
\,\left\{ \frac{{\bf I}_{N}}{N} + \alpha_1 \, \cone \, {\mathbf J}_{1}({\bfh}_1) +
\alpha_2 \,\left[ c_{2,1}
\, {\mathbf J}_{2}({\bfh}_2)   - c_{2,2} \, {\mathbf J}_{3}({\bfh}_3)
     - c_{2,3} \, {\mathbf J}_{4}({\bfh}_4)\right] \right\},
\end{align}

\noindent
where  ${\bf I}_{N}$ is the $N\times N$ identity matrix: $[{\bf I}_{N}]_{i,j}=1$ if $i=j$ and
$[{\bf I}_{N}]_{i,j}=0$ otherwise, and ${\mathbf J}_{q}({\bfh}_q), \; q=1,2,3,4$ are
\emph{network matrices} that are determined by the sampling pattern, the kernel function, and the
bandwidths. The  index $q$ defines the gradient network matrix for $q=1$, whereas the
values $q=2, 3, 4$ specify the curvature network matrices that correspond to the three terms
in ${\mcal{S}_2}(\xsam;\bfh_2)$ given by~\eqref{eq:S2}.
The elements of the network
matrices ${\mathbf J}_{q}(\bfh_{q})$ are given by the following equations

\begin{subequations}
\label{eq:network-mat}
\begin{align}
\label{eq:Jtilde}
[{\mathbf J}_{q}({\bfh}_q)]_{i,j} & = - \w(h_{q;i}) -\w(h_{q;j}) +
 [{\bf I}_{N}]_{i,j} \, \sum_{l=1}^{N} \left[ u_{i,l}(h_{q;i}) + u_{l,i}(h_{q;l})\right],
 \\
 \label{eq:J-weights}
\w(h_{q;i}) &   = \frac{K\left(  \frac{\bfs_{i} - \bfs_{j}}{h_{q,i}} \right)}
            {\sum_{i=1}^{N} \sum_{j=1}^{N} K\left(  \frac{\bfs_{i} - \bfs_{j}}{h_{q,i}} \right)},
            \quad q = 1, \ldots, 4.
\end{align}
\end{subequations}

The network matrices defined by~\eqref{eq:network-mat} are symmetric by construction.
It follows from~\eqref{eq:network-mat} that
the row and column sums vanish, i.e.,

\beq
\label{eq:sum-net-mat}
\sum_{j=1}^{N} [{\mathbf J}_{q}({\bfh}_q)]_{i,j} = 0.
\eeq

\noindent
Based on~\eqref{eq:Jtilde}, the diagonal elements are given by the following expression

\beq
\label{eq:diag-net-mat}
 [{\mathbf J}_{q}({\bfh}_q)]_{i,i} = \sum_{l=1,\neq i}^{N} \left[ u_{i,l}(h_{q;i}) + u_{l,i}(h_{q;l})\right].
\eeq

\noindent
Since the kernel weights are non-negative, it follows that the sub-matrices
${\mathbf J}_{q}({\bfh}_q)$ are \emph{diagonally dominant}, i.e.,
$\big| [{\mathbf J}_{q}({\bfh}_q)]_{i,i} \big| \ge \sum_{j \neq i}
\big|[{\mathbf J}_{q}({\bfh}_q)]_{i,j} \big|$.
It also follows from~\eqref{eq:prec-mat} and~\eqref{eq:sum-net-mat}  that

\begin{subequations}
\beq
\label{eq:sum-J}
\sum_{j=1}^{N} [{\mathbf J}(\bmthe)]_{i,j} = \frac{1}{N \lambda}.
\eeq
\end{subequations}


\subsection{Parameter inference}
\label{ssec:pi}

We have experimented both with maximum likelihood estimation and leave-one-out cross validation.
The former requires the calculation of the SLI partition function,
which is an  $O(N^3)$ operation for scattered data.
For large data sets the $O(N^3)$ complexity is a computational bottleneck.
Parameter inference by  optimization of a cross validation metric is computationally more efficient,
since  it is at worst an $O(N^2)$ operation as we show below.
The memory requirements for storing the precision matrix are $O(N^2)$ but can be significantly
reduced by using sparse matrix structures.
Let $\bftheta_{-\lambda}=(\alpha_{1},\alpha_{2},\mu,\mx)^{T}$
 represent the parameter vector excluding $\lambda$.
We use the following \emph{cross validation cost functional}

\beq
\label{eq:cost}
\Phi(\xsam;\bftheta_{-\lambda}) = \sum_{i=1}^{N}  | \hat{\Xo}_{i}(\bftheta_{-\lambda})  - \Xo_{i} \,|,
\eeq
where $\hat{\Xo}_{i}(\bftheta_{-\lambda})$ is the SLI prediction at  $\bfs_{i}$ based on the
reduced sampling set $\Sn - \{ \bfs_{i} \}$ using the parameter vector $\bftheta_{-\lambda}$
which applies to all $i=1, \ldots, N$.  The prediction
is based on the interpolation equation~\eqref{eq:sli-mode-pred-J} below and does not involve $\lambda$ (see discussion in Section~\ref{ssec:implem}).

The optimal parameter vector  excluding $\lambda$, i.e., $\bftheta_{-\lambda}$,
is determined by minimizing the cost functional~\eqref{eq:cost}:
\beq
\bftheta^{\ast}_{-\lambda} = \arg \min_{\bftheta_{-\lambda}} \Phi(\xsam;\bftheta_{-\lambda}).
\eeq

\noindent If $\tilde{H}(\xsam;\bftheta_{-\lambda})$ is  the energy estimated from~\eqref{eq:H-using-J}
and~\eqref{eq:prec-mat} by setting  $\lambda =1$, the optimal value $\lambda^{\ast}$ is obtained by minimizing
the negative log-likelihood with respect to $\lambda$  leading to the following solution (see~\ref{sec:app-A})

\beq
\label{eq:lambda}
\lambda^{\ast}    = \frac{2\tilde{H}(\xsam;\bftheta_{-\lambda})}{N}.
\eeq

We determine the minimum of the cross validation cost functional~\eqref{eq:cost} using the {\sc Matlab} constrained optimization
function \verb+fmincon+ with the \emph{interior-point} algorithm~\cite{Wright05}.
This function determines the local optimum nearest to the initial parameter vector.
We use initial guesses for  the parameters $\al_{1}, \al_{2}, \mu$, and we assume that
the parameters are constrained
between the lower bounds $[0.5, 0.5, 0.5 ]$ and the upper bounds $ [300, 300, 15 ]$.
 We investigated different initial guesses for the parameters which led to
 different local optima. We found, however, that the value of the cross  validation function
 is not very sensitive on the local optimum. In the $4D$ case study presented in Section~\ref{ssec:4D}, we also estimate 
 for comparison purposes  the global optimum using {\sc Matlab}'s global optimization tools.

\subsection{Predictive SLI model}
\label{ssec:SLI-prediction}
Let us now assume that the prediction point $\bfz_p$ is added to the sampling points.
To predict the unknown value of the field at  $\bfz_p$, we insert this point
in the energy functional~\eqref{eq:H-using-J}, which is then given by Eq.~\eqref{eq:H-pred-J} below.
Then, we determine the mode of the joint pdf~\eqref{eq:gibbspdf} with the prediction point
inserted in the energy functional. Thus, we  obtain a
\emph{mode prediction equation} for $\hXo_{p}$ given by~\eqref{eq:sli-mode-pred-J} below.

\subsubsection{Modification of kernel weights}
Upon inclusion of $\bfz_p$,
the weights~\eqref{eq:J-weights} of the network matrices~\eqref{eq:Jtilde}
 are modified as follows

\beq
\label{eq:weights-pred}
\w(h_{q;i})   = \frac{K\left(  \frac{\bfs_{i} - \bfs_{j}}{h_{q,i}} \right)}
{\sum_{i,j} K\left(  \frac{\bfs_{i} - \bfs_{j}}{h_{q,i}} \right)
+ \sum_{i} K\left(  \frac{\bfs_{i} - \bfs_{p}}{h_{q,i}}\right) +
\sum_{i} K\left(  \frac{\bfs_{i} - \bfs_{p}}{h_{q,p}}\right) },
\eeq

\noindent
where $\sum_{i,j} := \sum_{i=1}^{N} \sum_{j=1}^{N} $. The first term in the denominator concerns
interactions between sampling points. The second term involves local interactions between the sampling points
and the prediction point which result from inserting the prediction point in the local neighborhoods of
the sampling points, which control the bandwidths. Finally, the third term
also involves interactions between the prediction point and the sampling points,
but in this case the bandwidth is controlled by the former. Fig.~\ref{fig:points-pred} below illustrates the
difference between the second and third term in the context of the entire precision matrix.  The index $q$ is used to
distinguish between the weights linked to the gradient $(q=1)$ and the three weights $(q=2, 3, 4)$
linked to the curvature terms. The only difference between weights with different $q$ is the
bandwidth.  In the case of compactly supported kernels, different bandwidths imply that different
numbers of pairs are involved in the summations, since a pair separated by a distance that
exceeds the bandwidth does not contribute.
Calculation of the predictand  contributions in the denominator of~\eqref{eq:weights-pred}
is an operation with computational complexity $O(N)$ compared to $O(N^2)$  for the
interactions between sampling points. The latter term, however, is calculated once and
used for all the prediction points.

In addition to the weights that correspond to pairs of sampling points, there
are weights for combinations of sampling and prediction points, i.e.,

\beq
\label{eq:weights-pred-pi}
u_{p,j}(h_{q;p})   = \frac{K\left(  \frac{\bfs_{p} - \bfs_{j}}{h_{q,p}} \right)}
{\sum_{i,j} K\left(  \frac{\bfs_{i} - \bfs_{j}}{h_{q,i}} \right)
+ \sum_{i} K\left(  \frac{\bfs_{i} - \bfs_{p}}{h_{q,i}}\right) +
\sum_{i} K\left(  \frac{\bfs_{i} - \bfs_{p}}{h_{q,p}}\right) },
\eeq

\noindent
where $p=1, \ldots, P$, $j=1, \ldots, N$. The denominator of~\eqref{eq:weights-pred-pi} is identical
to that of~\eqref{eq:weights-pred}.

\subsubsection{SLI mode predictor}
Using the precision matrix formulation, the energy functional including the
predictand is given by

\begin{align}
\label{eq:H-pred-J}
{\hHx}(\xsam, \Xo_p ; \bftheta^{\ast}) = & \Hx(\xsam ; \bftheta^{\ast})  +
  J_{p,p}(\bftheta^{\ast})(\Xo_p - \mx)^2  + \sum_{i=1}^{N} (\Xo_{i} - \mx) \,J_{i,p}(\bftheta^{\ast})\,(\Xo_{p} - \mx)
  \nonumber \\
    &   +
  \sum_{i=1}^{N} (\Xo_{p} - \mx) \,J_{p,i}(\bftheta^{\ast})\,(\Xo_{i} - \mx).
\end{align}

The elements of the precision matrix that involve the prediction point are

\begin{subequations}
\label{eq:prec-Jp}
\begin{align}
\label{eq:prec-Jp-off}
[{\mathbf J}_{q}({\bfh}_q)]_{p,p} = & \sum_{i=1}^{N}   \left[ u_{i,p}(h_{q;i}) + u_{p,i}(h_{q;p})\right],
\\
\label{eq:prec-Jp-dia}
[{\mathbf J}_{q}({\bfh}_q)]_{i,p} = & -   \left[ u_{i,p}(h_{q;i}) + u_{p,i}(h_{q;p})\right], \; i \neq p.
\end{align}
\end{subequations}

\begin{figure}
        \centering
        \begin{subfigure}[b]{0.45\textwidth}
                \includegraphics[width=\textwidth]{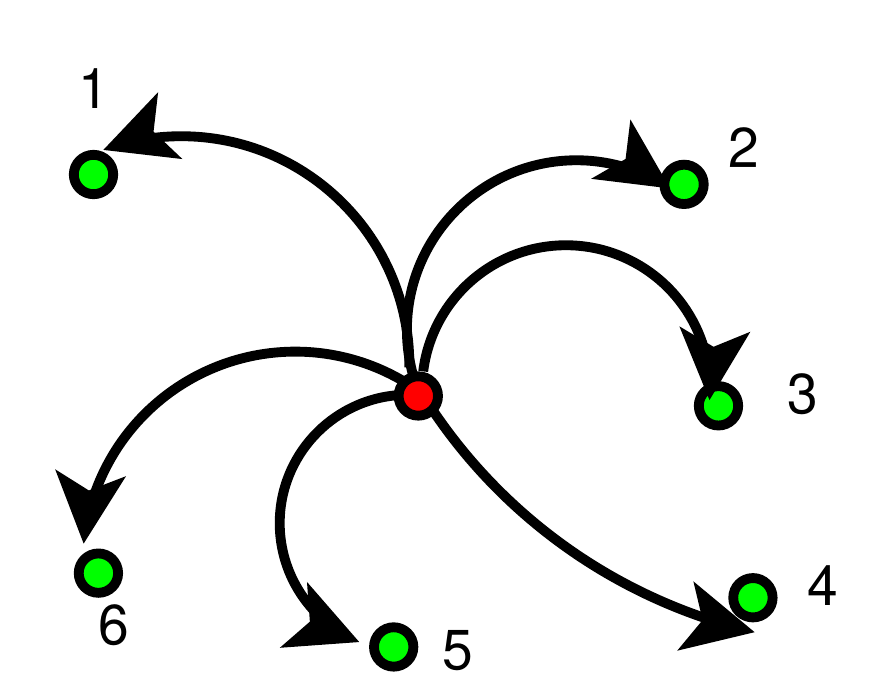}
                \caption{Bandwidth determined by $\bfz_p$. }
                \label{fig:prepoints1}
        \end{subfigure}%
        ~ 
        \begin{subfigure}[b]{0.4\textwidth}
                \includegraphics[width=\textwidth]{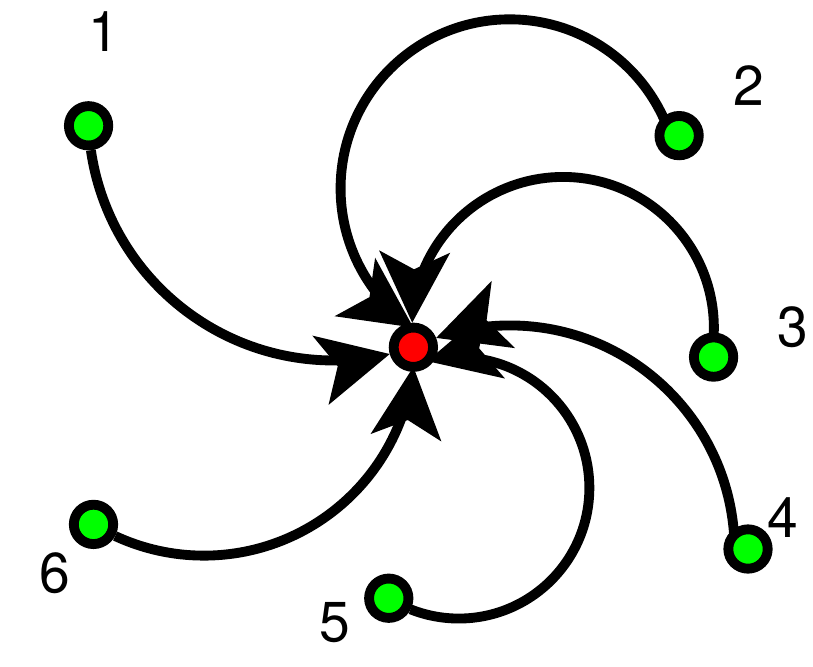}
                \caption{Bandwidth determined by $\bfs_{n}$. }
                \label{fig:prepoints2}
        \end{subfigure}
\caption{Schematic diagrams of terms
contributing to~\eqref{eq:H-pred-J} that include the prediction point $\bfz_p$ (center point) and
six sampling points $\bfs_{n}$ $(n=1,\ldots,6)$. The diagram on the left
(a) represents terms $J_{p,i}$ whereas
the diagram on the right (b) represents terms  $J_{i,p}$.
The point at the ``root'' of each arrow  determines the bandwidth for the
weight that involves the two points connected by the arrow.}
\label{fig:points-pred}
\end{figure}

Based on~\eqref{eq:prec-Jp-off} the  symmetry property
 $J_{p,i}(\bftheta^{\ast}) =J_{i,p}(\bftheta^{\ast})$ follows.  The coefficients
 $u_{i,p}(h_{q;i})$ and $u_{p,i}(h_{q;p})$  differ due to the
 different bandwidths used (in the former, the bandwidth is determined by the neighborhood
 of  the sampling point $\bfs_{i}$,  whereas in the latter by the neighborhood of $\bfz_{p}$.)
A schematic illustration of terms  in~\eqref{eq:H-pred-J} that involve
the predictand is
 given in Fig.~\ref{fig:points-pred}. The left diagram  corresponds to terms ``rooted'' at
 $\bfz_p$ (i.e., with coefficient $u_{p,i}(h_{q;p})$ that involves the bandwidth $h_p$),
 whereas the right hand side diagram  corresponds to
 terms rooted at the sampling points, i.e., with coefficients $u_{i,p}(h_{q;i})$.

The \emph{SLI mode predictor} is defined by the following equation

\beq
\label{eq:SLI-mode-pred}
\hXo_p = \arg \min_{\Xo_p} {\hHx}(\xsam, \Xo_p ; \bftheta^{\ast}),
\eeq

\noindent
where $ {\hHx}(\xsam, \Xo_p ; \bftheta^{\ast})$ is given by~\eqref{eq:H-pred-J}.
Minimization of the energy with respect to  $x_p$
 leads  to the following  {mode estimator}

\begin{align}
\label{eq:sli-mode-pred-J}
 \hXo_{p}   &   = \mx - \frac{\sum_{i=1}^{N}  \, \left[ J_{i,p}(\bftheta^{\ast})
+ J_{p,i}(\bftheta^{\ast}) \right] \, (\Xo_{i} - \mx)}{2\, J_{p,p}(\bftheta^{\ast})}
\nonumber \\
&   =
\mx - \frac{\sum_{i=1}^{N}  \,  J_{p,i}(\bftheta^{\ast}) \, (\Xo_{i} - \mx)}{ J_{p,p}(\bftheta^{\ast})},
\end{align}

\noindent
where the precision matrix elements are given by~\eqref{eq:Jtilde} using the modified kernel weights~\eqref{eq:weights-pred} and~\eqref{eq:weights-pred-pi}.

The SLI mode predictor can be generalized to $P$ prediction points as follows

\begin{subequations}
\label{eq:SLI-prediction}
\begin{equation}
\label{eq:H-mode-pred-J-multi}
  \bhXo_{p} = \bmx  -  \mathbf{\tilde{J}}_{P,S}(\bftheta^{\ast}) \, (\bXo - \bmx) ,
\end{equation}

\noindent
where $\mathbf{\tilde{J}}_{P,S}(\bftheta^{\ast})$ is a $ P \times N$ matrix given by

\begin{equation}
[\mathbf{\tilde{J}}_{P,S}(\bftheta^{\ast})]_{p,i} = J_{p,i}(\bftheta^{\ast})/ J_{p,p}(\bftheta^{\ast}).
\end{equation}
\end{subequations}

\subsubsection{Properties of SLI predictor}

The SLI prediction~\eqref{eq:SLI-prediction} is \emph{unbiased} in view of the vanishing row sum property~\eqref{eq:sum-J} satisfied by
 the network matrices and the precision matrix.
The SLI prediction~\eqref{eq:SLI-prediction}   is independent of
 the parameter $\lambda$ which sets the amplitude of the fluctuations, because
 the transfer matrix $\mathbf{\tilde{J}}_{P,S}(\bftheta^{\ast})$ is given by the ratio
 of precision matrix elements.
 This property is analogous to the independence of the
 kriging predictor from the random field variance. Hence,
  leave-one-out cross validation does not determine
 the optimal value of $\lambda$, which is obtained
 from~\eqref{eq:lambda}.

 The SLI predictor is not necessarily an exact interpolator. In particular, let us consider a point
$\bfs_{k}$, $k \in \{1, \ldots, N\}$, which is very close to $\bfz_p$. Based on~\eqref{eq:prec-Jp} and~\eqref{eq:sli-mode-pred-J},
$\hXo_{p} \rightarrow \Xo_k$ as $\bfz_p \rightarrow \bfs_k$ only if (i) $|u_{k,p}(h_p)| \gg |u_{i,p}(h_p)|$ and (ii)
$|u_{k,p}(h_k)| \gg |u_{i,p}(h_i)|$ for all $ i \neq k$.
Condition (i) materializes only for compactly supported kernels if $h_p \rightarrow 0$ which requires that the bandwidth be determined
by the nearest neighbor distance.   Condition (ii), on the other hand, requires that
$\|\bfs_{k} - \bfz_{p} \|/h_{k} \ll \|\bfs_{i} - \bfz_{p} \|/h_{i}$ for $i \neq k.$
This condition holds approximately at best if the sample is sparse around $\bfz_{p}$.

The computational complexity of the SLI predictor is $O(N^2 + P\, N)$.
The $O(N^2)$ term is due to the double summation over the sampling points
in~\eqref{eq:weights-pred}, which needs to be calculated only once.  The
remaining operations per each prediction point scale linearly with the
sample size, hence the $O(P\, N)$ dependence.
Based on the above, the dominant term (for fixed $P$) in the computational time  scales as $O(N^2)$.
In future work we will investigate
approximating the double summation in the denominator of~\eqref{eq:J-weights} and~\eqref{eq:weights-pred-pi}
with analytically evaluated double integrals over the kernel functions to  increase
the computational efficiency.

\section{Case Studies}
\label{sec:case}

We first consider two synthetic data sets, the first consisting of a time series and the
second of a four-dimensional test function. We then investigate a set of scattered real data
in two spatial dimensions.

\subsection{Time series with Mat\'{e}rn covariance function}
\begin{figure}
        \centering
        \begin{subfigure}[b]{0.49\textwidth}
                \includegraphics[width=\linewidth]{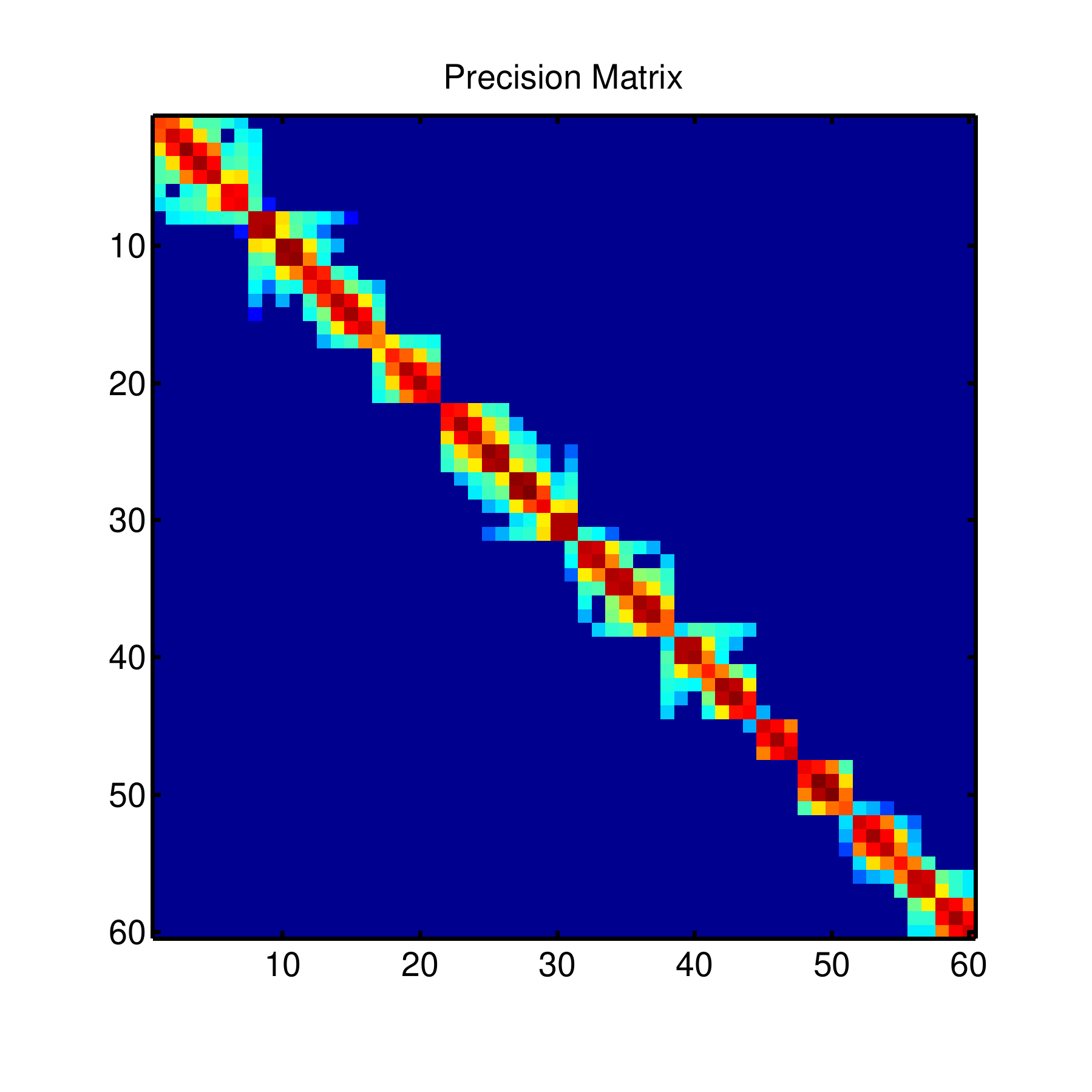}
                \caption{$\ln|J_{i,j}|$}
                \label{fig:lnabsPrec_1d}
        \end{subfigure}
        \begin{subfigure}[b]{0.49\textwidth}
                \includegraphics[width=\linewidth]{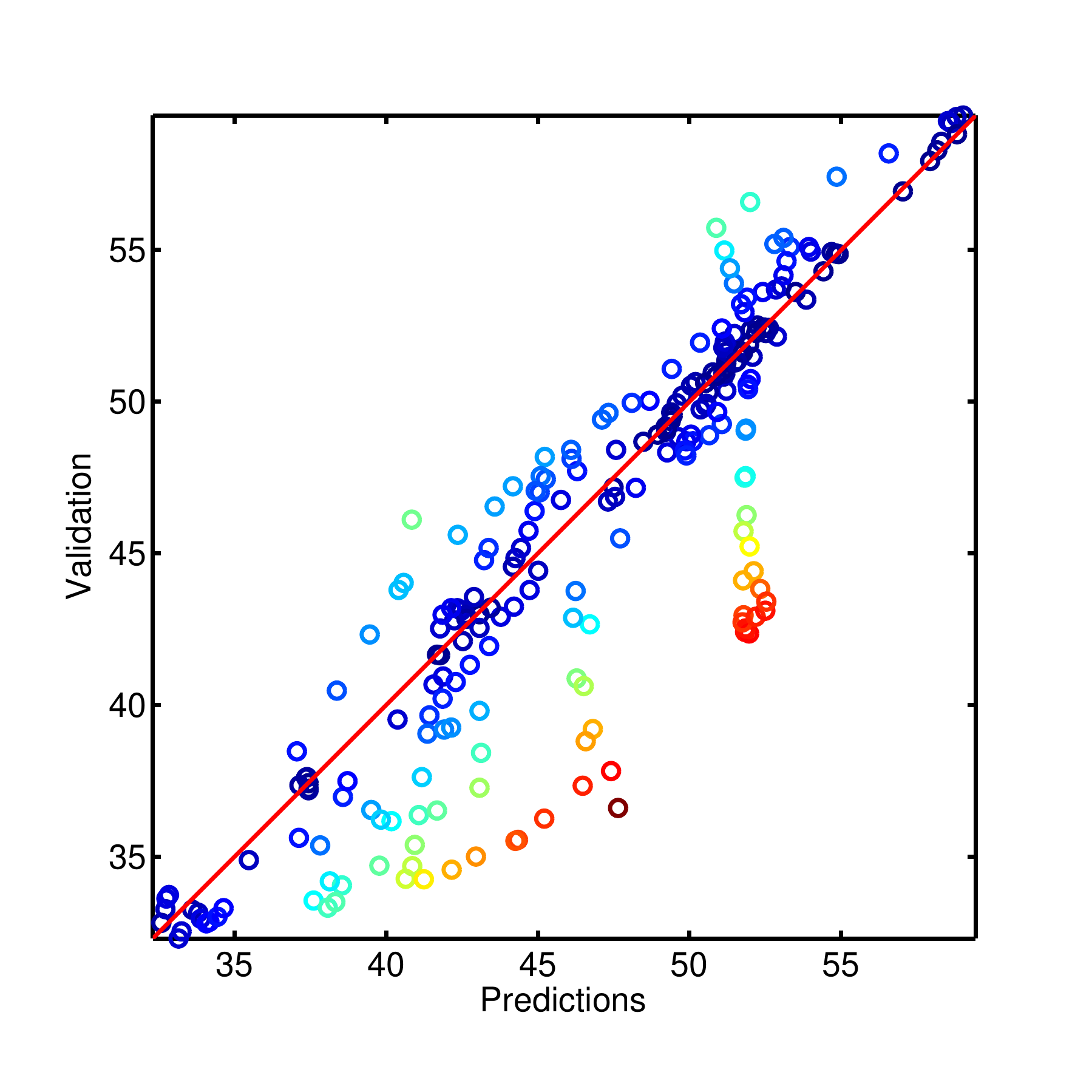}
                \caption{Predictions vs validation values}
                \label{fig:LVO_cv_1d}
        \end{subfigure}
\caption{\label{fig:SLI1d_Analysis}Analysis of time series with
Mat\'{e}rn correlations ($\sigma =10$, $\nu = 3.5$, $\xi = 10$).
(a) Logarithm of absolute value of the precision matrix; dark areas (blue online)
correspond to low values whereas lighter areas (green to red online) correspond to higher values.  (b)
Scatter plot of SLI predictions versus the respective values of the validation set. }
\end{figure}

We generate a time series of length $N=300$ from a
random process with Mat\'{e}rn covariance
$C(\tau) = \sigma^2 \, 2^{1-\nu} K_{\nu}(\tau/\xi) (\tau/\xi)^{\nu}/\Gamma(\nu)$,
where $K_{\nu}(\cdot)$ is the modified Bessel function of order $\nu$, $\Gamma(\cdot)$ is the gamma function,
 $\sigma =10$, $\nu = 3.5$ is the smoothness index, and $\xi = 10$ is the correlation time.
We use 60 randomly selected points as the training set (corresponding to an $80\%$ degree of thinning)
and the remaining
240 points as the validation set. The SLI optimal parameters using a quadratic kernel and  $k=2$ are given by
 $\alpha_{1} \approx 29.30, \alpha_{2} \approx 191.02, \mu \approx 1.11,
 \lambda  \approx  297.84$. The sparseness of the precision matrix is evident in
  Fig.~\ref{fig:lnabsPrec_1d}. The darkest areas correspond to negative infinity and reflect
distances for which the precision matrix vanishes.

\begin{figure}
  \centering
  \includegraphics[width=\linewidth]{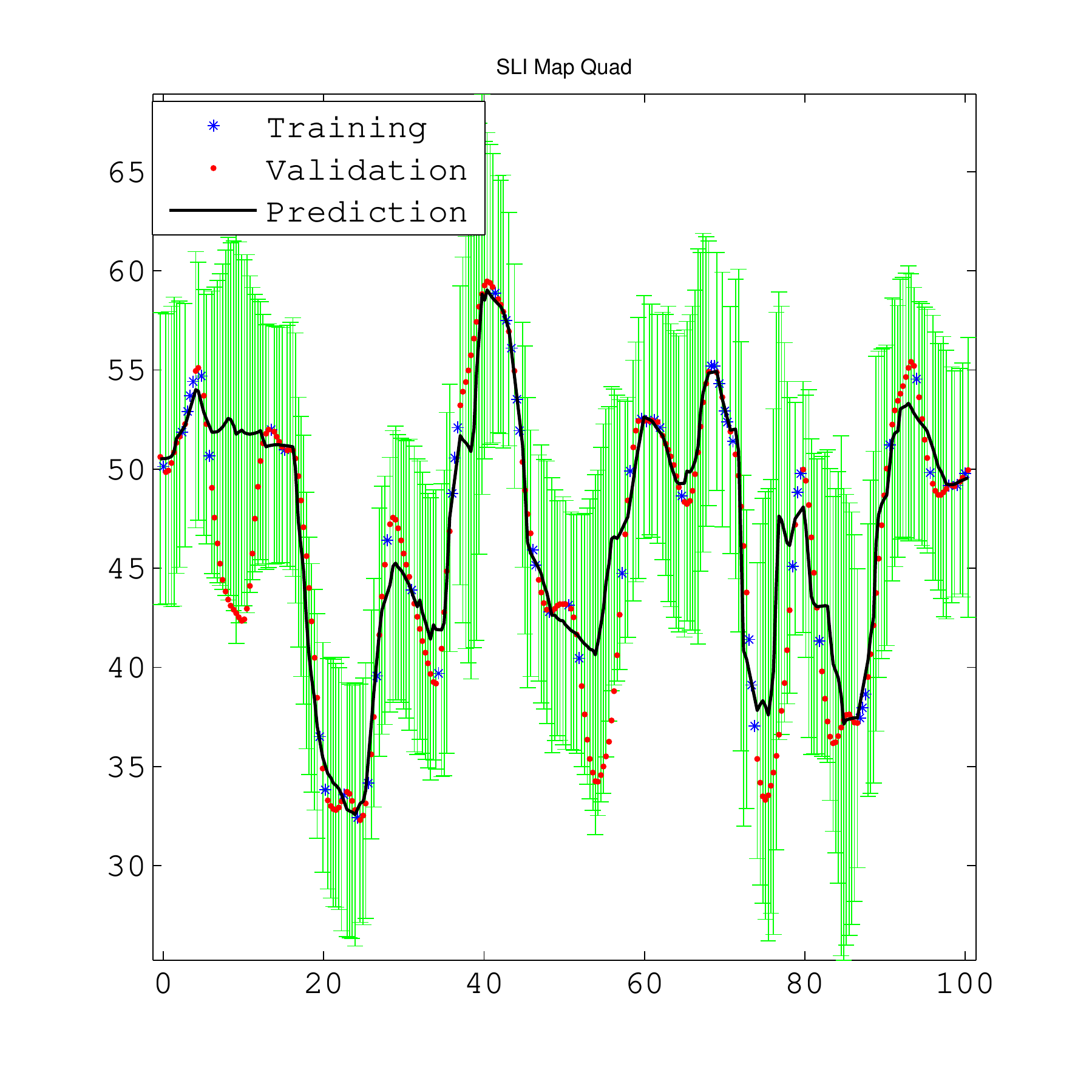}\\
  \caption{SLI predictions at 240 validation  points of time series with
  Mat\'{e}rn correlations ($\sigma =10$, $\nu = 3.5$, $\xi = 10$).
 The 60 training points are marked by stars (blue online), validation points are marked by dots (red online), and SLI predictions
  are marked by the continuous line (black online). Error bars are based on three times the conditional standard deviation  (green online). }\label{fig:SLI1d}
\end{figure}

  The prediction performance  is illustrated in
 the scatter plot of the SLI predictions versus the respective validation set values shown
 in Fig.~\ref{fig:LVO_cv_1d}. The Pearson correlation coefficient between the validation values and
 the predictions is $0.89.$ The splitting of the time series into training and validation
 sets is shown in Fig.~\ref{fig:SLI1d} along with the SLI predictions and associated error bars.
 The SLI predictions capture well general features of the time series. However, in areas
 of low sampling density the SLI predictions smoothes excessively the fluctuations
 in the original series.
 The SLI performance is excellent for the same degree of thinning, if the
 length of the initial time series increases to $3\,000$. On the other hand,
the prediction accuracy deteriorates for
 rougher random processes, such as a non-differentiable Mat\'{e}rn process with $\nu = 0.8$.

\subsection{Four-dimensional deterministic test function}
\label{ssec:4D}

\begin{figure}
        \centering
        \begin{subfigure}[b]{0.49\textwidth}
                \includegraphics[width=\linewidth]{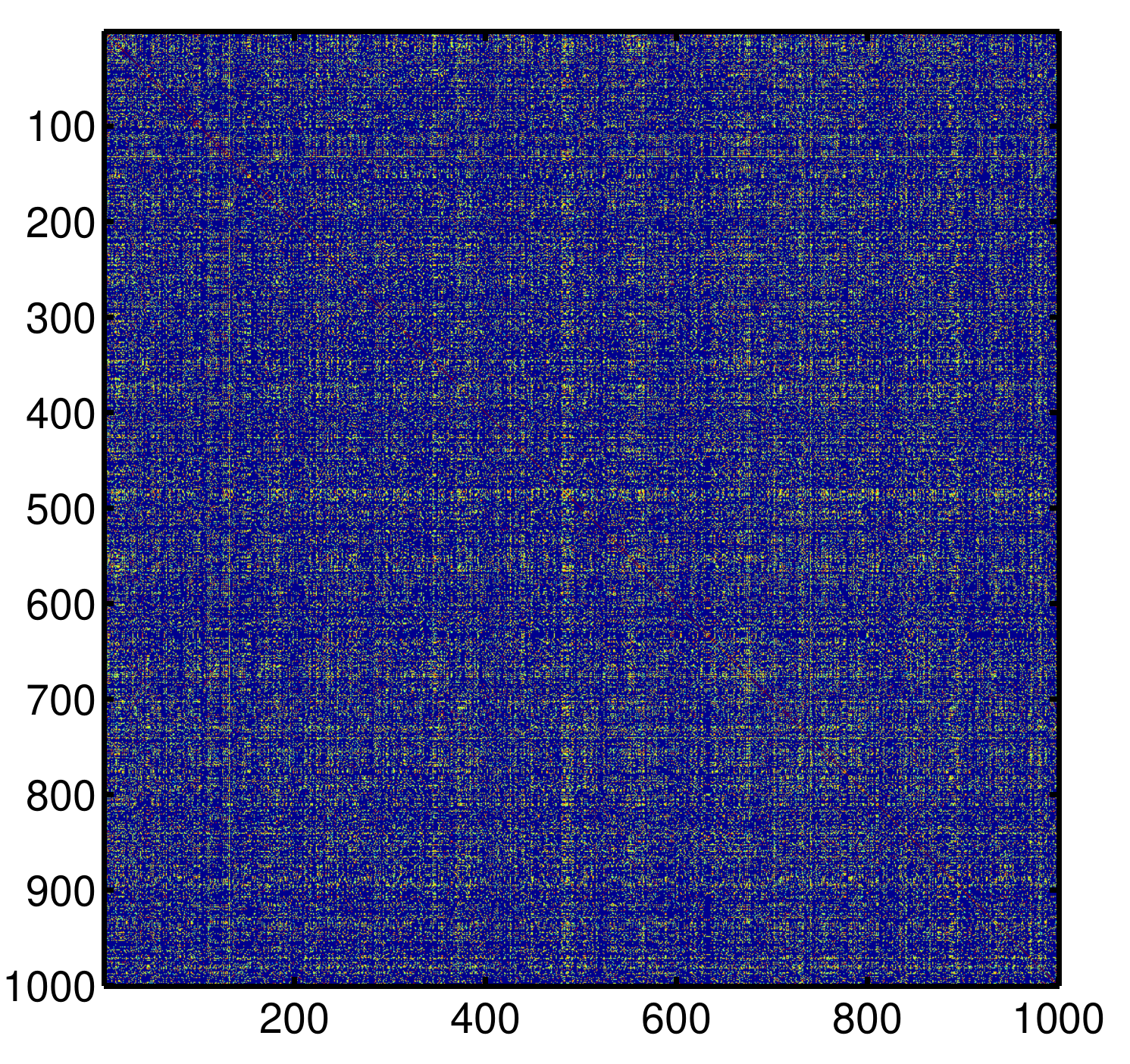}
                \caption{$\ln|J_{i,j}|$}
                \label{fig:lnabsPrec_4d}
        \end{subfigure}
        \begin{subfigure}[b]{0.49\textwidth}
                \includegraphics[width=\linewidth]{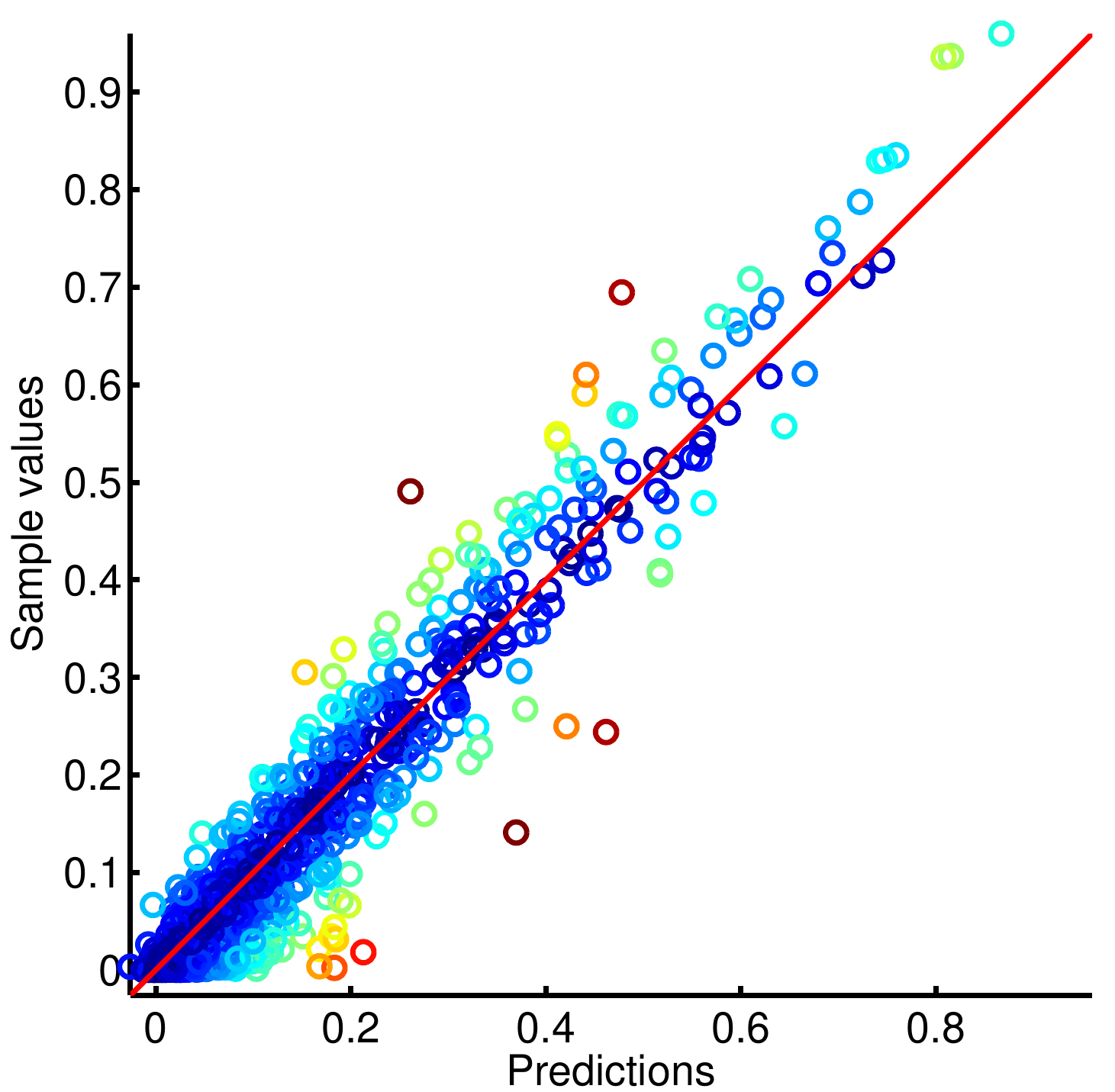}
                \caption{Predictions vs validation values}
                \label{fig:LVO_cv_4d}
        \end{subfigure}
\caption{\label{fig:SLI4d_Analysis} Analysis of the truncated exponential
function~\eqref{eq:trunc-expo}.
(a) Logarithm of the absolute value of the precision matrix; dark areas (blue online)
correspond to low values whereas lighter areas (green to red online) correspond to higher values.  (b)
Scatter plot of SLI predictions versus the respective values of the validation set. }
\end{figure}

We consider the  function $x(\bfs)$

\begin{equation}
\label{eq:trunc-expo}
x(\bfs) = A\, {\rm e}^{-2 \| \bfs - {\mathbf a}\| } \,
\prod_{i=1}^{4} s_{i} \,(1-s_{i}),
\end{equation}

\noindent
where $A = 500$ and
$\mathbf{a} = (0.3, 0.3, 0.3, 0.3)$,  defined over the
four-dimensional cube with unit length edges, i.e., for $\bfs \in [0, 1]^{4}$.
 We sample the function at $N=1\,000$ randomly selected points  over the unit cube, and
we generate a validation set of $N=1\,000$ points also by random selection.
The SLI optimal parameters for the quadratic kernel with  $k=2$ are given by
 $\alpha_{1} \approx  10.12, \alpha_{2} \approx 25.04, \mu \approx 1.64,
 \lambda  \approx 0.0193$ starting with initial values  $\alpha_{1}=10, \alpha_{2}=25, \mu=3$.
 Similar results in terms of cross validation performance are also obtained with different initial conditions that lead to
 different local optima.
  The cross validation measures for the parameters above are given by
 ME$=0.0046$,  MAE$=0.0320$, RMSE$=0.0459$, $r$$=0.96$.
 The sparse structure of the precision matrix is illustrated in Fig.~\ref{fig:lnabsPrec_4d}
 which displays the logarithm of the absolute value. The scatter plot of the validation values
 versus the respective SLI predictions is shown in Fig.~\ref{fig:LVO_cv_4d}  and demonstrates
 very good agreement at most points.

  We repeat the experiment by adding Gaussian noise to the sample. The standard deviation of the noise
 is set  to $\approx 10\%$ of  the maximum sampled value $x_{\max}$
  (in the simulations that we ran
 $x_{\max} \approx 1$). While the coefficients $\alpha_{1}$ and $\alpha_{2} $ remain practically unchanged, $\mu$ changes to $\approx 1.83$.
 The sparsity of the precision matrix is $\approx 76\%$.
   The respective cross validation measures are given by
 ME$=0.012$,  MAE$=0.047$, RMSE$=0.061$, and $r$$=0.93$.

 We also used the {\sc Matlab} global optimizer \verb+GlobalSearch+ with the same initial parameter vector as above to determine the SLI model parameters.
\verb+GlobalSearch+ uses a scatter-search algorithm to generate starting points (initial parameter guesses). The minimization is
conducted using \verb+fmincon+ to determine the local minimum close to the current starting point.
We use the  lower and upper bounds defined in Section~\ref{ssec:pi} to
constrain the space of the starting points. \verb+GlobalSearch+ investigates a set of 66 starting points, and convergence to a local
minimum is achieved for all of them.
 The globally optimal SLI parameters are estimated as $\alpha_{1} \approx  1.50, \alpha_{2} \approx 224.62, \mu \approx 2.01,
 \lambda  \approx 0.748$.
 The respective validation measures are given by ME$=0.0055$,  MAE$=0.045$,  RMSE$=0.063$, and    $r=0.93$. These measures do not differ significantly
 from those obtained with the locally optimum solution.   The MAE value is lower for the global optimum, which is expected since MAE reflects the value
 of the cost function~\eqref{eq:cost}. On the other hand, the RMSE obtained with the global optimum is slightly higher than that of the local optimum.
  This result indicates that quite different  parameter vectors can lead to similar cross validation results.
  This behavior has also been observed with covariance models whose parameter vector involves more than the variance
  and the correlation length~\cite{zuk09,dth11}.

\subsection{Radioactivity data in two dimensions}

This example focuses on  daily means of radioactivity gamma
dose rates over part of the Federal Republic of Germany. The data
were provided by the German automatic radioactivity monitoring
network for the Spatial Interpolation Comparison Exercise 2004 (SIC
2004)~\cite{dubois05}. This data set is well studied and
thus allows easy comparisons with other methods~\cite{dthetal08}.
The $1\,008$ stations are partitioned into a
\textit{training set} of $200$ randomly selected locations and a
\textit{validation set} of $808$ locations where predictions are
compared with the observations.  Two different scenarios are
investigated: A \textit{normal} data set corresponding to typical
background radioactivity measurements, and an \textit{emergency}
data set, in which a local release of radioactivity in the southwest corner
of the monitored area was simulated using a dispersion process  to
obtain a few values with magnitudes around  10 times higher than the
background. The rates are measured in nanoSievert per hour (nSv/h).
The normal training set follows the Gaussian distribution with the
minimum around 58 nSv/h and the maximum around 153~nSv/h. In  the
emergency training set there are two values $>1\,000$~nSv/h,
with the maximum at $1\,499$~nSv/h.
We compare the prediction performance of the SLI model against the $808$ values of the
validation set.

\subsubsection{Normal data}
\begin{figure}
        \centering
        \begin{subfigure}[b]{0.49\textwidth}
                \includegraphics[width=0.8\linewidth]{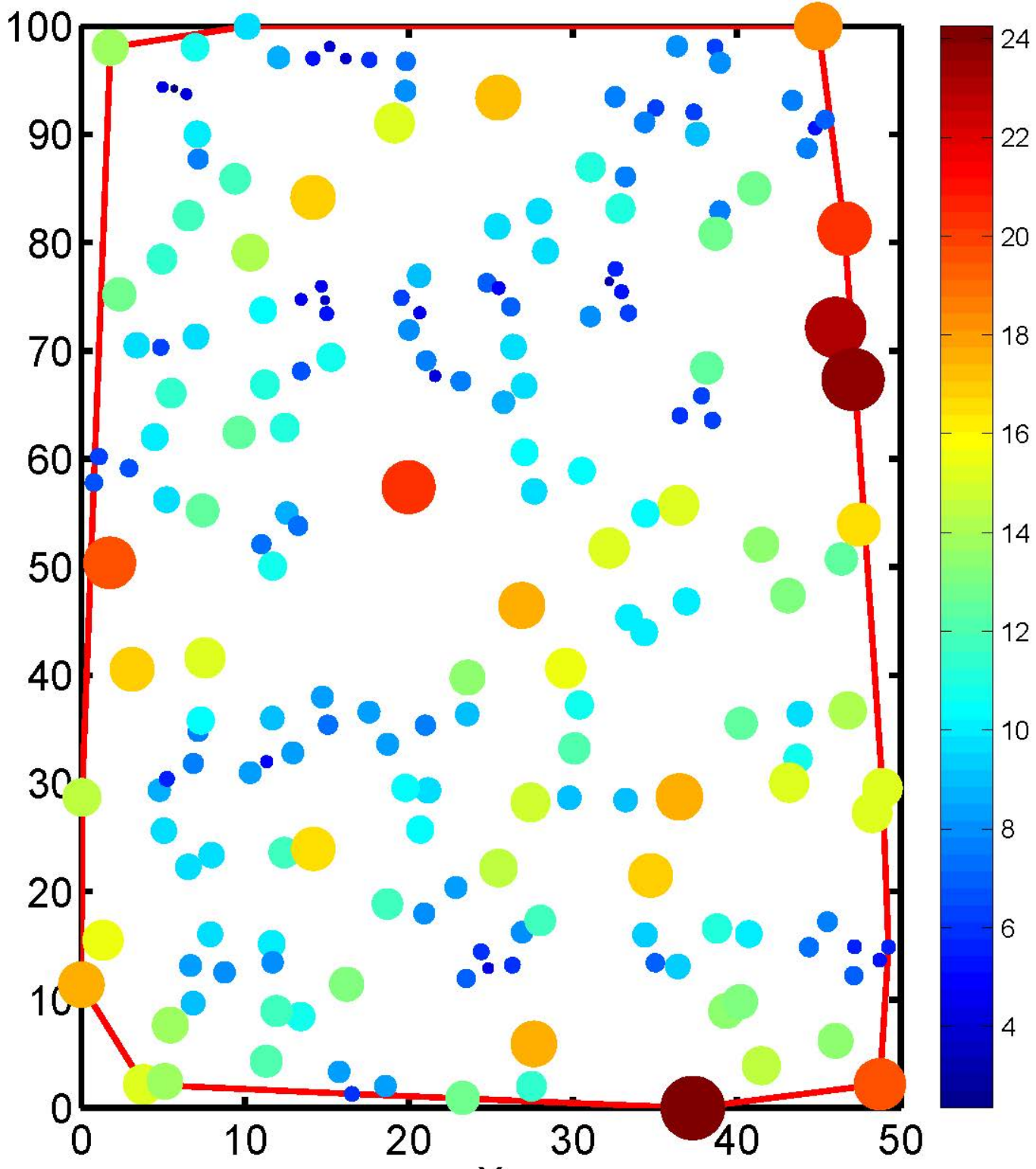}
                \caption{Relative bandwidth}
                \label{fig:rel-band}
        \end{subfigure}
        \begin{subfigure}[b]{0.49\textwidth}
                \includegraphics[width=0.9\linewidth]{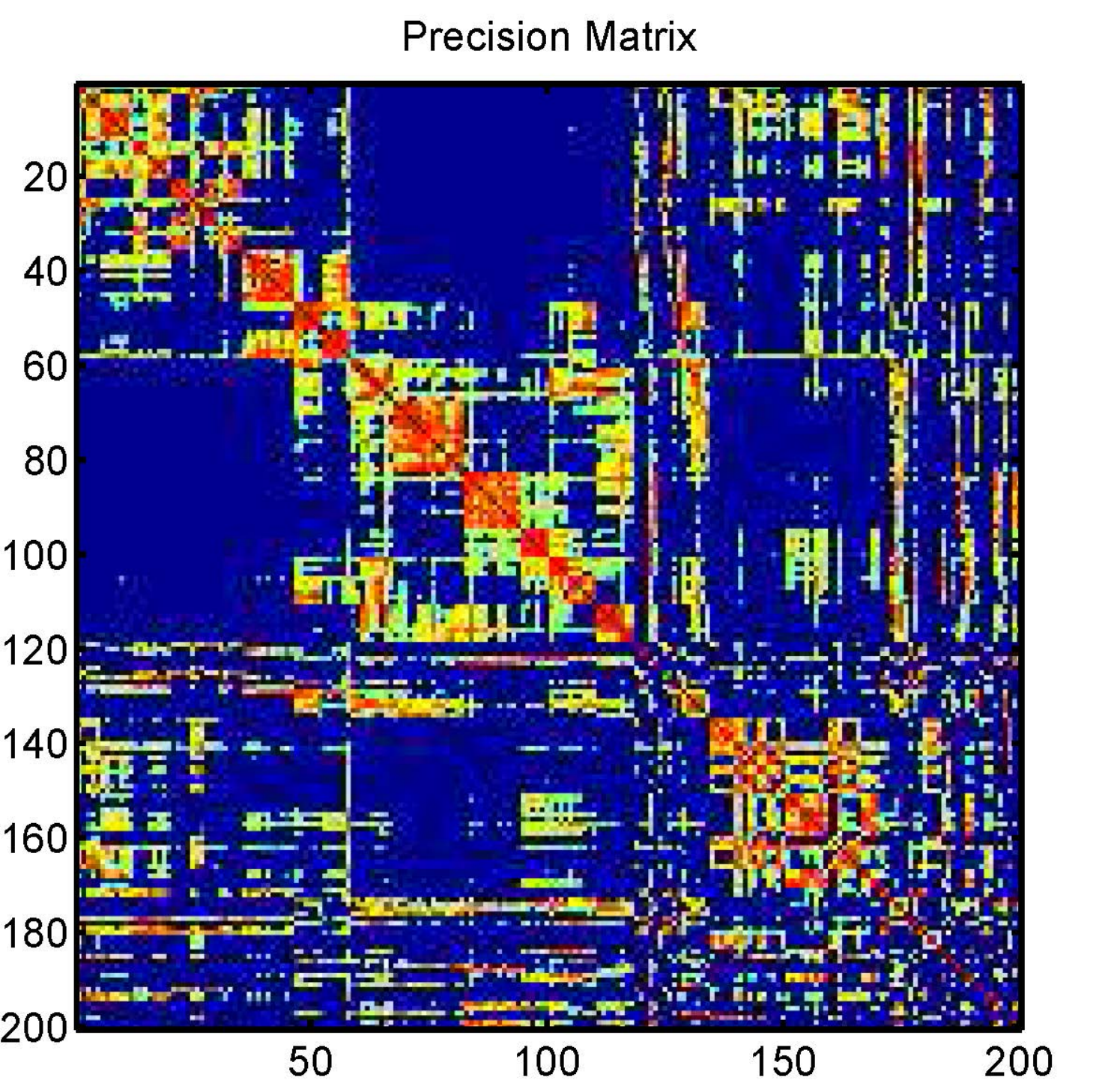}
                \caption{$\ln|J_{i,j}|$}
                \label{fig:lnabsPrec}
        \end{subfigure}
\caption{Analysis of SIC 2004 normal data set using normalized coordinates;
 the longest side is set to 100  and the aspect ratio is maintained.
(a) Bubble plot of the relative size of local bandwidths: larger circles correspond to
bigger bandwidths.  The continuous line along the domain boundary marks the
convex hull of the sampling set. (b)
Logarithm of the absolute values of the precision matrix elements. Darker areas (blue online) correspond to
lower values, whereas lighter areas (red online) correspond to higher values. }
\end{figure}

For normal data, the optimal SLI parameters
based on the  training set with a quadratic kernel  and  $k=2$ are given by
 $\alpha_{1} \approx 143, \alpha_{2} \approx 47.56, \mu \approx 2.64,
 \lambda  \approx 3.24\times 10^3$.
Figure~\ref{fig:rel-band} illustrates the relative  values of the bandwidths used.
Higher values correspond to more isolated points in areas of low
sampling density and along the boundaries of the convex hull of the domain.
Figure~\ref{fig:lnabsPrec} presents the natural logarithm of the absolute value of the
precision matrix.  Overall, about $ 32 \%$ (i.e., 12\,718) of
 the total number of pairs  yield nonzero precision values, implying that the sparsity of the
 precision matrix is $\approx 68\%$.

The cross-validation results are tabulated in Table~\ref{tab:cv-sic04-normal}.
The cross validation   measures (based on the validation set) obtained in a recent study by means of \emph{Ordinary Kriging} are: ME$=-1.36$,
 MAE$=9.29$, RMSE$=12.59$, $r$$=0.78$~\cite{dthetal08}.
 These values are in close agreement with the SLI results in Table~\ref{tab:cv-sic04-normal}.

\begin{figure}
        \centering
        \begin{subfigure}[b]{0.5\textwidth}
                \includegraphics[width=\linewidth]{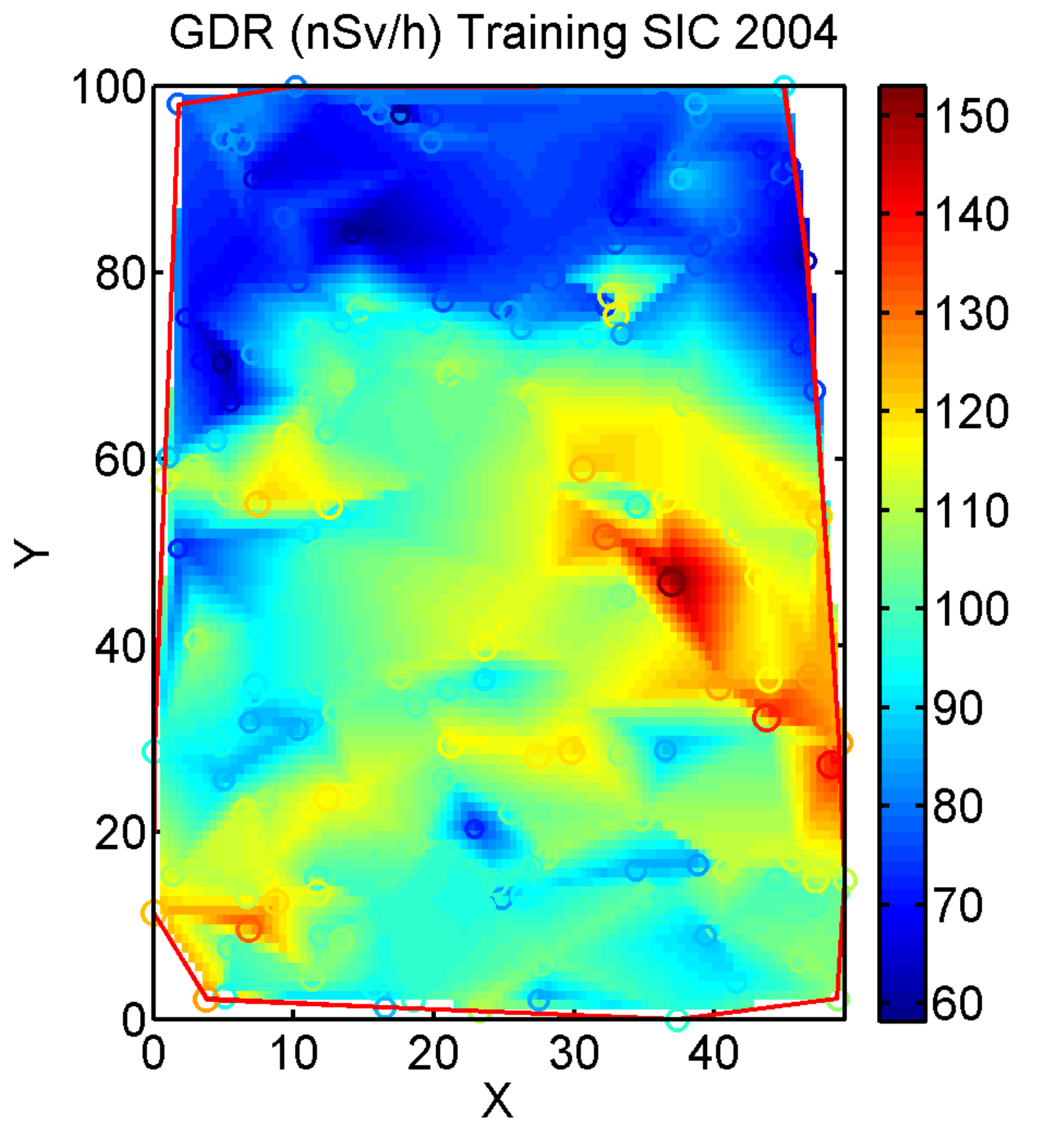}
                \caption{Bilinear interpolation}
                \label{fig:bil-interp}
        \end{subfigure}%
        ~ 
        \begin{subfigure}[b]{0.5\textwidth}
                \includegraphics[width=\linewidth]{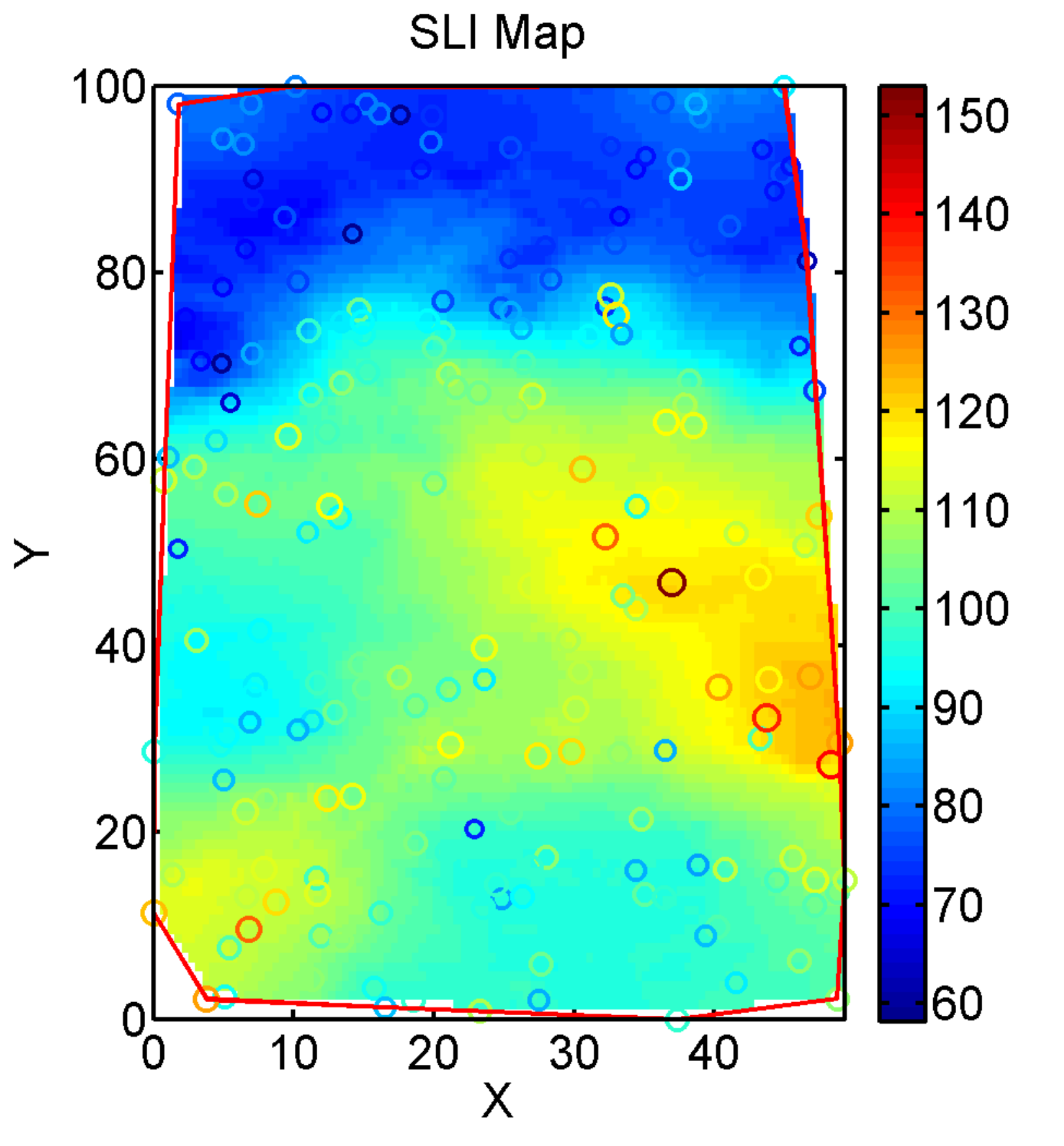}
                \caption{SLI interpolation}
                \label{fig:sli-interp}
        \end{subfigure}
\caption{\label{fig:map} Map of  background radioactivity rates in Germany (based on
 200 training data)
using a mapping grid with $100$ nodes per side.
(a) Bilinear interpolation using the \emph{griddata} command of  {\sc Matlab}. (b)
SLI interpolation map using the optimal parameters reported in the text.}
\end{figure}

Various geostatistical and machine learning methods have been applied to the
  SIC 2004 data (neural networks, geostatistics, and splines). Excluding the results of some
  poor performers, the cross validation measures obtained are in the following ranges~\cite{dubois05}:
  {ME} $\in [-1.39 , -0.04]$ and $\in [0.20, 1.60]$,  {MAE} $\in [9.05 ,  12.10]$,
{RMSE} $\in [12.43 ,  15.90]$, and $r$ $\in [ 0.64 ,  0.79]$. Hence, the SLI cross validation results
 are close to the best performers. Fig.~\ref{fig:map}  presents a map of the radioactivity pattern generated by
SLI and contrasts it with the map generated by bilinear interpolation. The SLI spatial pattern is smoother and
thus appears more realistic. Its smoothing effect near the sample values, however, is more pronounced
than that caused by bilinear interpolation.

\begin{table}[h]
\centering
\caption{\label{tab:cv-sic04-normal} Cross validation performance measures for SIC 2004 normal data. The second row presents the performance of
the SLI predictor at the  808 validation set  points.
{ME:} Mean error (bias); {MAE:} Mean absolute error; {MARE:} Mean absolute relative error; {RMSE:}
Root mean square error; {$r$:} Pearson correlation
coefficient.}
\begin{tabular}{lccccc}
\hline
{{SLI}}&{ME}&{MAE}&{MARE}&{RMSE}&{$r$}\\\hline
{Validation set}&$-$1.30&9.30&0.09&12.62&0.78\\\hline
\end{tabular}
\end{table}

\begin{figure}
  \centering
  \includegraphics[width=0.7\textwidth]{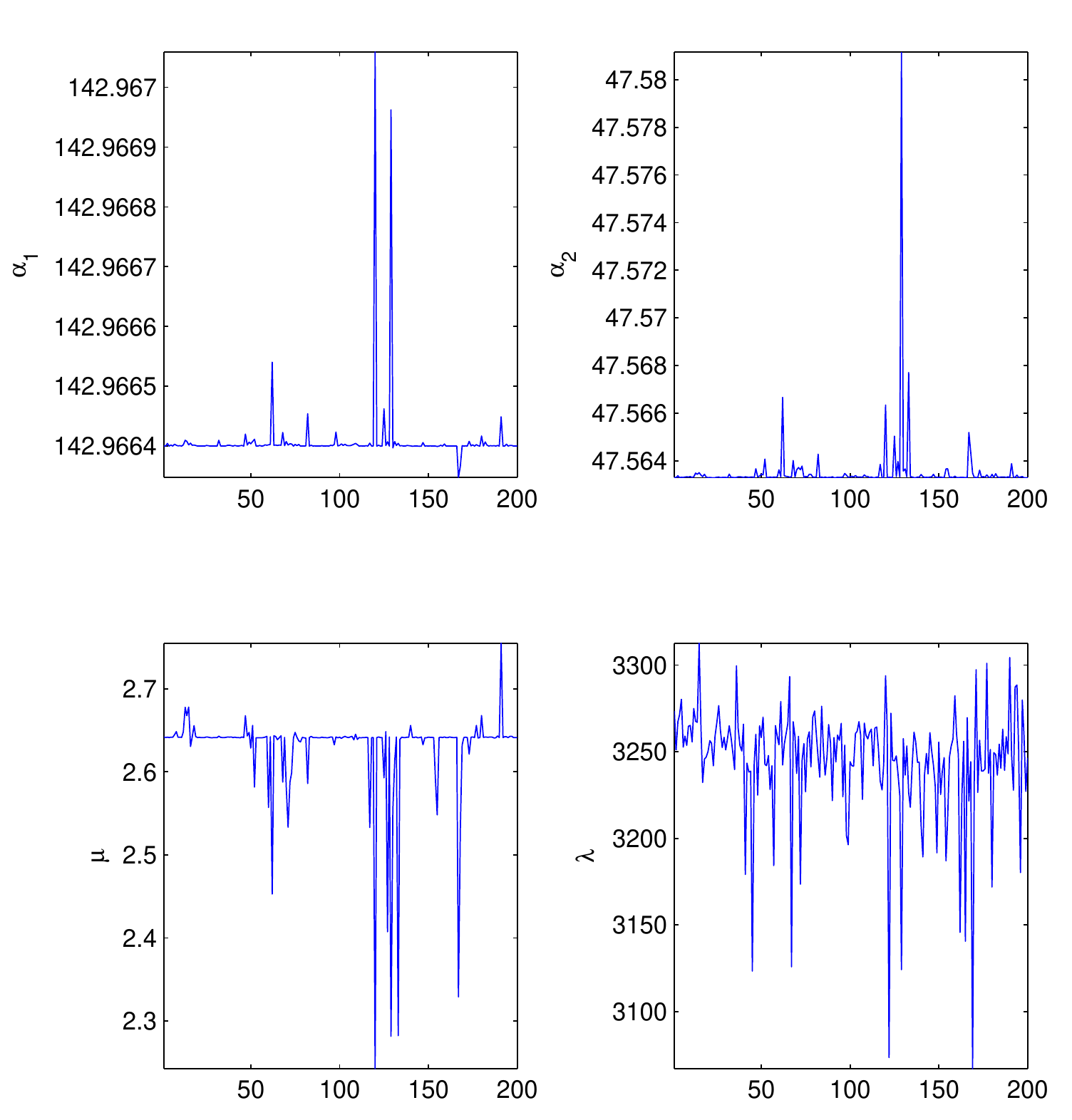}\\
  \caption{Variation of SLI parameters estimated by removing one value at a time
  from the 200 training locations of the SIC 2004 normal data.}\label{fig:LVO-params}
\end{figure}

We also conduct a stability analysis by
removing one sampling point at a time and determining the optimal SLI model using
leave-one-out cross validation with
that point removed.
The variation of the SLI parameters is shown in Fig.~\ref{fig:LVO-params};
  $\alpha_1, \alpha_2$ and $\mu$ are quite stable, whereas $\lambda$ shows
 more variability.
 The spikes in the plots of Fig.~\ref{fig:LVO-params} are exaggerated by using a narrow vertical range
 to better illustrate the parameter variability.  For $\alpha_1$, $\alpha_2$ the maximum relative
 variation (with respect to the mean) ranges from a fraction of a thousandth (for $\alpha_1$) to
 few thousandths  (for $\alpha_2$); $\mu$ shows stronger variations, whereas the strongest variation
 is exhibited by $\lambda$, since the latter is a scaling factor that determines
 the overall energy of the ensemble of points and compensates for variations in the other parameters.
 We believe that the parameter variations exhibited in Fig.~\ref{fig:LVO-params} are, at least partially
 due to extremely slow variation of the cost function over a region of the parameter space, a condition
 also observed in maximum likelihood estimation of spatial models with Mat\'{e}rn
 covariance~\cite{zuk09}.  This slow variation implies \emph{quasi-degeneracy} of the parameter vector; the
 quasi-degeneracy implies
 that vectors which are very far in parameter space may lead
 to very similar cost function values.
 More recently, the difficulties involved in nonlinear fits of multi-parametric models to data have been investigated in~\cite{Transtrum10}.

\subsubsection{Emergency data}
For the SIC 2004 emergency data, cross validation results with different kernel
functions (tricubic, exponential, and quadratic) are shown in Table~\ref{tab:cv-sic04-joker}.
The SLI parameters are initialized using the optimal values for the normal data.
The last row of the table is based on estimation with a quadratic kernel
function  after removing the three highest values.
The best results in Table~\ref{tab:cv-sic04-joker}
are obtained with the quadratic kernel including all the data.
The optimal SLI parameters are $\alpha_{1} \approx 143, \alpha_{2} \approx 47.56, \mu \approx 2.69,
 \lambda  \approx 4.32\times 10^5$. The parameters, except for $\lambda$, are close to their
  normal case counterparts. The difference in $\lambda$ is due to the much higher variance
 of the emergency data set.

The variation of the SLI parameters in leave-one-out cross validation
exhibits similar patterns as for the normal data, except that more pronounced
 variations of $\lambda$ are observed when the extreme values are removed.
 The  precision matrix has
 23\,232 non-zero elements, implying a sparsity of $\approx 42\%$, in contrast with 12\,718 (sparsity $\approx 32\%$)
 in the normal case. This difference clearly illustrates the
 dependence of the precision matrix on the sample values in addition to the sampling pattern.
SLI does not rely on
estimating the variogram function, and thus it is not hindered by the presence of extreme values.
 On the other hand, geostatistical methods rely on the variogram function,
 which may not be reliably estimated in such cases~\cite{Giralids05}.
 The Pearson correlation coefficient is
significantly lower than in the normal set due to underestimation of the
extreme values, while
the Spearman rank correlation coefficient is comparable to the normal case.
The cross validation measures obtained in SIC 2004
are in the following intervals~\cite{dubois05}:
{ME} $\in [-11.10 , -0.12]$ and $\in [0.41, 19.71]$,  {MAE} $\in [14.85 ,  146.36]$,
{RMSE} $\in [45.46 ,  212.10]$, and $r$ $\in [ 0.02 ,  0.86]$.
The best performance in terms of both MAE and RMSE was obtained by means of a
Generalized Regression Neural Network~\cite{Timonin06}.
Looking at the scatter plot of MAE versus RMSE ---Fig.~6 in~\cite{dubois98}--- the
SLI performance is closer to the geostatistical and spline methods.

\begin{table}[h]
\centering
\caption{\label{tab:cv-sic04-joker} SLI cross validation performance measures for SIC 2004 emergency data.
The second row presents the performance of
the SLI predictor at the  808 validation set  points.
The first five cross validation measures are as described in the caption of
Table~\ref{tab:cv-sic04-normal}, and $r_S$ is the Spearman correlation coefficient.}
\begin{tabular}{lcccccc}
\hline
{{SLI} Kernel function}&{ME}&{MAE}&{MARE}&{RMSE}&{$r$}& $r_S$\\\hline
{Tricubic}: $(1-u^3)^3$ &5.78&24.22&0.20&81.33&0.25&0.57\\\hline
{Exponential}: $ {\mathrm e}^{-u}$ &6.06&23.84&0.19&79.78&0.34&0.63\\\hline
{Quadratic}&3.04&23.16&0.17&75.63&0.43 &0.77\\\hline
{Quadratic (outliers removed)}&$-$8.28&16.46&0.10&81.41&0.27&0.77 \\\hline
\end{tabular}
\end{table}

\section{Discussion}
\label{sec:disc}

\subsection{Connections with machine learning}

The SLI model is similar to $k$-Nearest Neighbors (KNN), since both methods employ
an optimal neighborhood range. In the case of KNN a uniform optimal number of nearest neighbors is
determined, and the estimate at an unmeasured point is simply the mean of its $k$ nearest neighbors.
In SLI, a locally optimal neighborhood size is determined implying that the number of neighbors used in
prediction varies locally. In addition, the estimate is a weighted mean of the neighbor values, in which the
weights are determined by the kernel function and the bandwidths. In this respect, SLI is similar to the Nadaraya-Watson kernel
regression method~\cite{Nadaraya64,Watson64} and to the Support Vector Machine algorithm~\cite{Vapnik00}.
SLI can also be viewed as a particular type of Gaussian process with a  sparse inverse covariance kernel, which
could be used as an alternative to the sparse Gaussian process framework to improve the computational efficiency
of  predictions~\cite{Csato02}.

In this study we  formulated the SLI model using the spatial locations $\Sn$ as inputs and the respective
values of the scalar field values as outputs.  This framework is appropriate for scattered spatial data.
It is possible, however, to use more general input variables instead of the  spatial locations,
so long as a suitable measure of distance can be defined.

\subsection{Notes on implementation }
\label{ssec:implem}
We presented a ``plain vanilla'' version of the SLI model. Modifications that can increase the
flexibility but also the complexity of the model are possible.  The local kernel bandwidths are determined
by fixing the neighbor order $k$ and using a uniform scaling parameter $\mu$.  Alternatively, one
can consider estimating $k$ from the data and using a locally varying $\mu$. With respect to the latter,
potential gains should be weighted against the loss of computational efficiency that will result from
the significant increase of the parameter vector size.
While our estimate of  $\mx$ is based on the
sample mean, it is possible to estimate $\mx$ by means of the leave-one-out cross validation procedure.
It is also possible to replace $\mx$ with a space-dependent trend function.

The present version of the SLI model does not involve anisotropy.
Nevertheless, anisotropy is important in cases such as
 the radioactivity emergency data~\cite{Spil11}: the best
performing method in SIC 2004 for this set
was a general regression neural network with an anisotropic
Gaussian kernel function. Similarly, in SLI it is possible to use weighted Euclidean distances or
Minkowski metrics instead of the classical Euclidean distance~\cite{Atkeson97}.
SLI can also be extended to spherical surfaces,
a case which is relevant for global geospatial data.
In addition, the SLI model can  capture correlations in
 higher-dimensional, abstract feature spaces equipped with a suitable distance.

At this point there is no rigorous physical interpretation of the
coefficients $\alpha_{1}$, $\alpha_{2}$ and $\lambda$.
In general, higher values of $\alpha_{1}$ ($\alpha_{2}$) imply higher cost for gradient (curvature),
whereas $\lambda$ controls the overall ``energy''.
In the continuum case (i.e., for Spartan random fields) coefficients $\alpha_1$ and $\alpha_2$ are related
to a rigidity coefficient and a characteristic length~\cite{dth03b,dth11}.
A similar correspondence can also be established for data distributed on rectangular grids.
In contrast, such relations are not available for scattered data.
 Even in the continuum and grid cases, however, statistical measures
such as the variance  and the correlation length have a nonlinear dependence
on the SSRF model parameters~\cite{dth03b,dth11}.
A reasonable initial value for $\mu$ is around $2$--$3$,
to allow even compactly supported kernel functions to build local neighborhoods containing at least a few
data points. For $\alpha_1$ and $\alpha_2$,
we have used positive values between the arbitrary bounds of 0.5 and 300. Exploratory runs with different initial conditions
can help to locate a reasonable starting point. Alternatively, a global optimization approach can be used as in Section~\ref{ssec:4D}.

We have opted for a cross-validation cost functional which is based on the mean absolute error.
It is possible to use different cost functionals that involve a linear combination of
validation measures such as the mean absolute error and the root mean square error.
Most results for the case studies investigated above were obtained using
an interior-point optimization method that searches for local minima of the cost function.
In all of the cases that we have investigated (including data not presented herein), the local optimization
led to reasonable  cross validation measures which were comparable to those obtained with other methods.
As we have shown in the case of $4D$ synthetic data,  searching
for global optima does not necessarily lead to significant performance improvement.
The investigation  of global optimization methods with different data sets, however, deserves further attention.

\section{Conclusions}
\label{sec:concl}
The  SLI model presented above provides a bridge between geostatistics and machine learning.
It is  based on an exponential joint density  which involves an energy functional with an explicit
precision (inverse covariance) matrix.
The latter is constructed by superimposing network sub-matrices that implement
local interactions between neighboring field
values in terms of kernel functions.
The algorithmic complexity of SLI missing value estimation
scales linearly with the sample size except for a global $O(N^2)$ term which
is, however, computed once for all the prediction points.
Hence,  the leave-one-out cross-validation approach can be
used to efficiently infer the SLI model parameters.

For missing data on rectangular grids (ongoing research) the computational complexity of the
SLI method can be simplified to  linear scaling with $N$,
because $\mcal{S}_{1}$  and $\mcal{S}_{2}$
can be calculated without kernel functions~\cite{zuk13,zuk13b}.
In addition,  calculating and storing the large $N\times N$  distance matrix is not necessary
in this case.
In conclusion, the SLI model is a promising tool for the analysis of big spatial data.
In future research we will investigate the extension of the model to space-time data.
Finally, the {\sc Matlab} code used for the case studies in Section~\ref{sec:case}
 is available at the  web address of the Geostatistics laboratory:
 \url{http://www.geostatistics.tuc.gr/4940.html}.

\section*{Acknowledgment}
The research presented in this manuscript was funded by the project SPARTA 1591: ``Development of
Space-Time Random Fields based on Local Interaction Models and Applications in the Processing of
Spatiotemporal Datasets''. The project SPARTA is implemented under the ``ARISTEIA'' Action of the
 operational programme ``Education and Lifelong Learning'' and is co-funded by the European Social Fund
 and National Resources.

\section*{References}
\bibliographystyle{Chicago}


\begin{appendix}

\section{Minimization of NLL}
\label{sec:app-A}

For the pdf given
by~(\ref{eq:gibbspdf}), the log-likelihood is given by
\begin{equation}
\label{loglik}
{\mathrm {LL}}(\xsam ;\bftheta) \doteq \ln{L(\xsam ;\bftheta)} = -\Hx(\xsam ;\bftheta) -
\ln{Z(\bftheta)}.
\end{equation}
The partition function  in~(\ref{loglik})  is given by the multiple integral
\begin{equation}
\label{eq:Z} Z(\bftheta) = \prod_{i=1}^{N} \,
\int_{-\infty}^{\infty} d\Xo_i \, \exp\left( -\Hx(\xsam ;\bftheta) \right).
\end{equation}
The square gradient and square curvature terms do not depend on $\mx$ because they involve differences $x_{i}-x_{j}$.
Hence, we can express~\eqref{eq:H-using-J} as follows
\begin{equation}
\label{eq:en-est} \Hx (\xsam ;\bftheta)  =
\frac{1}{2}
  \xsam ^{T} \, {\mathbf J}(\bftheta) \,\xsam + \frac{\mx^2}{\lambda } - \frac{2\mx \, \smx}{\lambda},
\end{equation}
where $\smx$ is the sample mean.
Maximizing the ${\mathrm {NLL}}$ with respect to $\mx$, using~\eqref{eq:en-est} for the
energy functional, yields
\[
\mx = \smx.
\]

Since this fixes the parameter $\mx$, we can use expression~\eqref{eq:H-using-J} for the energy functional.
We apply the scaling transformation $\Hx(\xsam;\bftheta) = \tilde{H}(\xsam;\bftheta_{-\lambda})/\lambda$,
where $\bftheta_{-\lambda}$ is the parameter vector except for $\lambda$
and $\tilde{H}(\xsam;\bftheta_{-\lambda})$ is $\lambda$-independent.
The transformation $H(\cdot) \mapsto \tilde{H}(\cdot)$ is equivalent to
$\Xo_{i} \mapsto y_{i} = (\Xo_{i} -\smx)/\sqrt{\lambda}$.
Let us then define the scaled partition function $\tilde{Z}(\bftheta_{-\lambda})$ by
means of
\begin{align}
\label{eq:Z1} \tilde{Z}(\bftheta_{-\lambda}) = & \prod_{i=1}^{N} \,
\int_{-\infty}^{\infty} dy_i \, \exp\left( -\tilde{H}({\mathbf y};\bftheta_{-\lambda}) \right)
\nonumber \\
   = &   \lambda^{-N/2} \prod_{i=1}^{N} \,
\int_{-\infty}^{\infty} d\Xo_i \, \exp\left( -{H}(\xsam;\bftheta) \right)
\nonumber \\
    =   & \lambda^{-N/2} \,{Z}(\bftheta) .
\end{align}

In light of the above transformations, the dependence of ${\mathrm {NLL}}$ on  $\lambda$ takes the following explicit form
\[
{\mathrm {NLL}}(\xsam;\bftheta)
=\frac{\tilde{H}(\xsam;\bftheta_{-\lambda})}{\lambda} + \frac{N}{2} \ln\lambda +
 \ln\tilde{Z}(\bftheta_{-\lambda}).
\]

Hence, by minimizing NLL with respect to $\lambda$, i.e.,
 $\frac{d {\mathrm {NLL}}(\xsam;\bftheta)}{d\lambda} =   0$,  we obtain
the following expression for the optimal $\lambda$:
\begin{align}
\lambda^{\ast}    = \frac{2\tilde{H}(\xsam;\bftheta_{-\lambda})}{N}.
\end{align}

From the Gaussian joint pdf~\eqref{eq:H-using-J} it follows that
\[
{\tilde{Z}(\bftheta_{-\lambda})} = (2\pi)^{N/2}\, \left\{\det\left[\tilde{J}(\bftheta_{-\lambda})\right] \right\}^{-1/2},
\]
where $\tilde{J}(\bftheta_{-\lambda}) = \lambda \, {J}(\bftheta)$.
We insert the optimal value $\lambda^{\ast}$ in ${\mathrm {NLL}}$ and use the
expression above for the log-partition function which leads to
\begin{align}
\label{eq:NLL}
{\mathrm {NLL}}(\xsam;\bftheta_{-\lambda}) =  &  \frac{N}{2} \ln \left( \frac{2\tilde{H}(\xsam;\bftheta_{-\lambda})}{N} \right)
 + \frac{N}{2}\ln(2\pi) -  \frac{1}{2}\det\left[\tilde{J}(\bftheta_{-\lambda})\right].
\end{align}
The NLL~\eqref{eq:NLL} is minimized numerically
using the {\sc Matlab} constrained minimization
function \verb+fmincon+. Constraints are used to ensure that the parameter
values are positive. The  log-determinant
is calculated numerically using the singular value decomposition of the precision matrix.
This is a procedure with numerical complexity $O(N^3)$ for a full rank matrix. For this reason,  we
 use cross validation instead of maximum likelihood for parameter inference.

\end{appendix}

\end{document}